\documentclass[letterpaper, 10 pt, conference]{ieeeconf}  

\IEEEoverridecommandlockouts                              

\usepackage{cite}
\usepackage{amsmath,amssymb,amsfonts}
\usepackage{algorithmic}
\usepackage{graphicx}
\usepackage{subfigure}
\usepackage{textcomp}
\usepackage{xcolor}
\usepackage{multirow}
\usepackage{capt-of} 

\usepackage{booktabs}

\def\aboveCaptionFig{0pt}
\def\belowCaptionFig{0pt}

\usepackage{flushend}
\usepackage{cite}
\makeatletter
\let\NAT@parse\undefined
\makeatother
\usepackage[colorlinks,linkcolor=blue,anchorcolor=blue,citecolor=blue,urlcolor=blue,hyperfootnotes=true]{hyperref}   
\usepackage[all]{hypcap} 
\usepackage{placeins}

\title{\LARGE \bf
Learning to Initialize Trajectory Optimization for Vision-Based Autonomous Flight in Unknown Environments
}

\author{Yicheng Chen$^{1,2}$, Jinjie Li$^{2}$, Wenyuan Qin$^{3}$, Yongzhao Hua$^{3*}$, Xiwang Dong$^{1,3,4}$, and Qingdong Li$^{1}$
}

\begin{document}

\twocolumn[{%
\renewcommand\twocolumn[1][]{#1}%
\maketitle
\thispagestyle{empty}
\pagestyle{empty}
\begin{center}
    \setlength{\abovecaptionskip}{0pt} 
    \setlength{\belowcaptionskip}{0pt}
    \centering
    \includegraphics[width=1\linewidth]{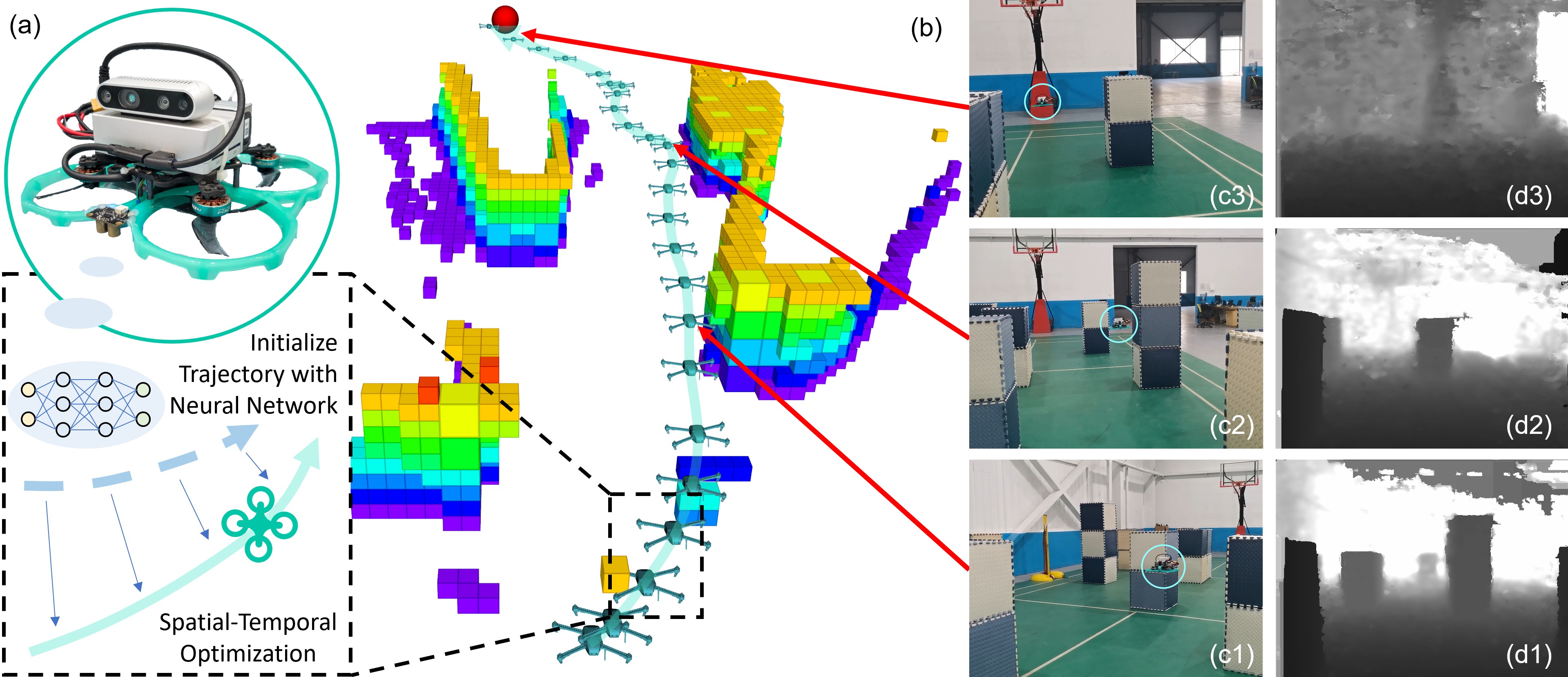}
    \captionof{figure}{\textbf{A real-world demonstration of fully autonomous flight using the proposed Neural-Enhanced Trajectory Planner (NEO-Planner).} The drone has no prior knowledge of the environment, and the entire software stack runs onboard in real-time.  (a) Our quadrotor platform. (b) The map built during the flight. (c1)-(c3) Three snapshots during the flight. (d1)-(d3) The corresponding first-person-view depth images.}
    \label{exp_process}
\end{center}%
}]

\begingroup
\renewcommand{\thefootnote}{}
\footnotetext{This work was supported by the National Natural Science Foundation of China under Grants U2241217, 62473027, 62473029, 62403038, and 62203032, and the Beijing Natural Science Foundation under Grants JQ23019 and 4232046.}
\footnotetext{$^{1}$ School of Automation Science and Electrical Engineering, Beihang University, Beijing 100191, P.R. China}
\footnotetext{$^{2}$ Department of Mechanical Engineering, The University of Tokyo, Tokyo 113-8656, Japan.}
\footnotetext{$^{3}$ School of Artificial Intelligence (Institute of Artificial Intelligence), Beihang University, Beijing 100191, P.R. China.}
\footnotetext{$^{4}$ Institute of Unmanned System, Beihang University, Beijing 100191, P.R. China.}

\footnotetext{* Corresponding author}
\footnotetext{Emails: yicheng@buaa.edu.cn, jinjie-li@dragon.t.u-tokyo.ac.jp, wyqin@buaa.edu.cn, yongzhaohua@buaa.edu.cn, xwdong@buaa.edu.cn, liqingdong@buaa.edu.cn}
\endgroup

\begin{abstract}
Autonomous flight in unknown environments requires precise spatial and temporal trajectory planning, often involving computationally expensive nonconvex optimization prone to local optima. To overcome these challenges, we present the Neural-Enhanced Trajectory Planner (NEO-Planner), a novel approach that leverages a Neural Network (NN) Planner to provide informed initial values for trajectory optimization. The NN-Planner is trained on a dataset generated by an expert planner using batch sampling, capturing multimodal trajectory solutions. It learns to predict spatial and temporal parameters for trajectories directly from raw sensor observations. NEO-Planner starts optimization from these predictions, accelerating computation speed while maintaining explainability. Furthermore, we introduce a robust online replanning framework that accommodates planning latency for smooth trajectory tracking. Extensive simulations demonstrate that NEO-Planner reduces optimization iterations by 20\%, leading to a 26\% decrease in computation time compared with pure optimization-based methods. It maintains trajectory quality comparable to baseline approaches and generalizes well to unseen environments. Real-world experiments validate its effectiveness for autonomous drone navigation in cluttered, unknown environments.

\end{abstract}

\textbf{Code:} \href{https://github.com/Amos-Chen98/neo-planner}{https://github.com/Amos-Chen98/neo-planner}

\textbf{Video:} \href{https://youtu.be/UoroRe-euDk}{https://youtu.be/UoroRe-euDk}

\section{Introduction}
Spatial-temporal motion planning aims to generate collision-free trajectories with refinement in both energy and time. This has been a challenging problem for autonomous drones in unknown environments because it is required to precisely handle the complexity from both the environment and the drone dynamics, while ensuring a real-time performance for high-frequency replanning. 

Optimization-based approaches \cite{tordesillas_mader_2022, zhou_swarm_2022, wang_geometrically_2022} formulate motion planning as an optimization problem, incorporating various constraints and costs into a single objective function. However, due to the high-dimensional variables and intricate constraints, the optimization problem is inherently nonconvex, often leading to convergence at local optima. Additionally, computational efficiency heavily depends on the quality of the initial values used for optimization \cite{ichnowski_deep_2020, yoon_learning-based_2023}.

A common strategy for providing initial values is to leverage path planners—such as A* \cite{hart_formal_1968} or Rapidly-exploring Random Tree (RRT) \cite{lavalle_randomized_2001}—to generate sparse waypoints for trajectory optimization. However, these planners ignore drone dynamics. They often produce unrealistic guesses and incur runtimes too long for high-frequency replanning \cite{gao_obstacle-aware_2023}. One alternative is sampling multiple initial configurations and optimizing each. This approach raises computational cost linearly with the number of samples, making it impractical for micro aerial vehicles with limited onboard computing resources \cite{tordesillas_panther_2022}. Therefore, an efficient, low-cost method is needed to produce reasonable initial trajectories.

On the other hand, learning-based methods have gained significant attention because neural networks can model complex nonlinear mappings \cite{willard_integrating_2023} and execute inference rapidly \cite{brunke_safe_2022}. Researchers have applied both supervised and reinforcement learning to tasks such as obstacle avoidance \cite{tordesillas_deep-panther_2023, song_learning_2023, sanket_ajna_2023}, wilderness navigation \cite{loquercio_learning_2021}, formation flight \cite{shi_neural-swarm2_2022}, and autonomous racing \cite{song_reaching_2023, kaufmann_champion-level_2023}. For example, Tordesillas et al. \cite{tordesillas_deep-panther_2023} propose a perception-aware planner, but it assumes perfect knowledge of obstacle trajectories, which limits real-world applicability. Song et al. \cite{song_learning_2023} demonstrate hardware-in-the-loop simulations yet require an abrupt takeover by a state-based controller if the network fails. Loquercio et al. \cite{loquercio_learning_2021} achieve high-speed flight, but their method struggles to maintain temporal consistency over long horizons, reducing success rates. The key challenge remains: how to generate reliable, explainable spatial–temporal trajectories that harness the advantages of learning-based approaches.

To overcome these limitations, we propose a learning-based approach to initialize trajectories and refine them through optimization. Several studies have combined learning and optimization for trajectory planning. For instance, Tang et al. \cite{tang_learning_2018} generate an initial path with a neural network and refine it via quadratic programming—but only for obstacle-free fly-to-target tasks. Banerjee et al. \cite{banerjee_learning-based_2020} warm-start sequential convex programming for obstacle avoidance but assume fully known obstacle models. Marino et al. \cite{marino_multi-uavs_2024} use a distributed two-branch network to predict both trajectories and collision constraints from point clouds, yet validation is confined to idealized simulations with perfect sensing. Other related work includes \cite{ichnowski_deep_2020, yoon_learning-based_2023, xue_combining_2024, wu_deep_2024}.

In contrast to previous studies that integrate learning into optimization, our research introduces two key innovations. First, our neural network directly generates trajectories from raw sensor observations, enabling seamless deployment in real-world scenarios. Second, the network outputs both spatial and temporal parameters to initialize trajectory optimization. Our simulations indicate that this approach is more effective than providing only spatial parameters. The main contributions of this paper are as follows:

\begin{itemize}
    \item 
    Neural-Enhanced Trajectory Planner (NEO-Planner): A novel approach that integrates a Neural Network Planner into an optimization-based framework, reducing planning costs while generating explainable, high-quality trajectories.
    \item 
    Robust Online Trajectory Replanning Framework: An online replanning framework that enables autonomous flight in unknown environments with tolerance to planning latency.
    \item
    Extensive Validation: A series of simulations and real-world experiments demonstrating the efficiency of our approach compared to existing methods and confirming its feasibility for real-world applications.
\end{itemize}

\section{Methodology}

\begin{figure*}[t]
    \setlength{\abovecaptionskip}{-3pt} 
    \setlength{\belowcaptionskip}{\belowCaptionFig}
    \centerline{\includegraphics[width=1\linewidth]{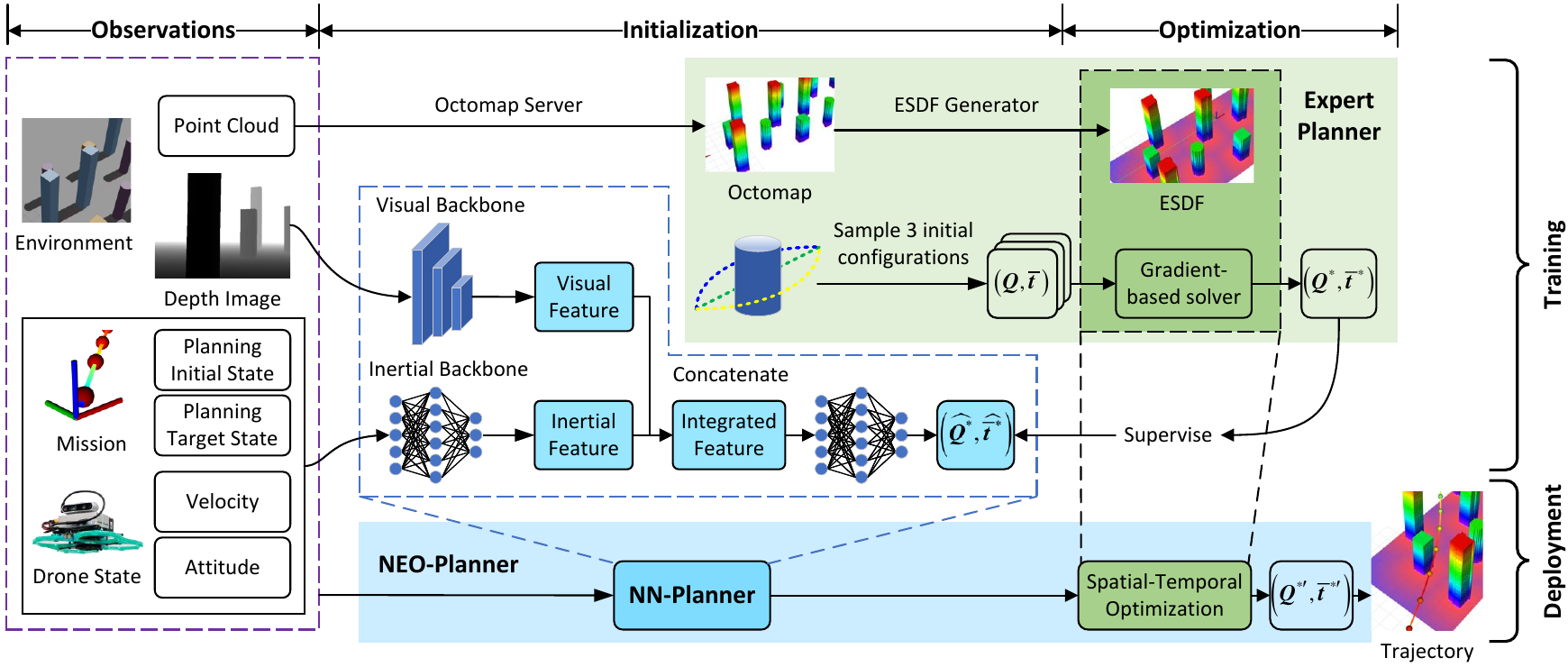}}
    \caption{\textbf{System overview}. The NEO-Planner leverages a neural network to generate high-quality initial trajectories from onboard observations and subsequently conducts spatial-temporal optimization on the neural network's output. The NN-Planner is trained using supervised learning, with training cases provided by an expert planner based on a standard mapping-planning-control stack.\vspace{-2mm}}
    \label{system_overview}
\end{figure*}

\subsection{System Overview}

Fig. \ref{system_overview} shows an overview of the system. We first parameterize the trajectories using MINCO representation (Section \ref{MINCO}), where a polynomial trajectory is completely defined by the waypoints and time allocation $(\boldsymbol{Q}, \overline{\boldsymbol{t}})$. The NEO-Planner performs spatial ($\boldsymbol{Q}$)-temporal ($\overline{\boldsymbol{t}}$) optimization (Section \ref{optimization}) on the trajectories with initial values generated by a NN-Planner (Section \ref{nn_planner}). The NN-Planner is trained using supervised learning. We leverage a powerful yet computationally-expensive expert planner (Section \ref{expert}) to provide the training data. In addition, we inroduce an online replanning framework in Section \ref{online_framework}.

\subsection{Trajectory Parameterization}
\label{MINCO}
We denote scalars in standard $x \in \mathbb{R}$ or $X\in \mathbb{R}$, vectors in bold lowercase $\boldsymbol{x} \in \mathbb{R}^n$, and matrices in bold uppercase $\boldsymbol{X} \in \mathbb{R}^{n \times m}$. In addition, we denote the time variable in $t$, time points in $t_i$, time intervals in $\overline{t}_i:=t_i-t_{i-1}$, and then $t_i=\sum_{j=1}^i \overline{t}_j$. The coordinate system contains the world frame $\{\mathcal{W}\}$ (ENU: X East, Y North, Z Up) and the body frame $\{\mathcal{B}\}$ (FLU: X Forward, Y Left, Z Up). The geometric variables are in the world frame if not specifically annotated.

We parameterize the trajectory using the MINCO representation \cite{wang_geometrically_2022}. Compared to alternatives such as Bernstein polynomials or B-splines, MINCO decouples spatial shape from timing, giving our neural network a direct interface to predict both profiles and thereby enabling higher-quality trajectory initializations. In flat output space \cite{mellinger_minimum_2011}, we represent a $D$-dimensional $M$-piece $t$-indexed polynomial trajectory $\boldsymbol{p}(t)$ as
\begin{equation}
\begin{gathered}
\mathfrak{T}_{\mathrm{MINCO}}=\left\{\boldsymbol{p}(t):[0, t_M] \to  \mathbb{R}^D \mid \boldsymbol{C}=\mathcal{C}(\boldsymbol{Q}, \overline{\boldsymbol{t}}),\right. \\
\left.\boldsymbol{Q} \in \mathbb{R}^{D \times (M-1)}, \overline{\boldsymbol{t}} \in \mathbb{R}_{>0}^M\right\},
\end{gathered}
\end{equation}
where $\boldsymbol{C}$ are trajectory coefficients,
$\boldsymbol{Q}=\left[\boldsymbol{q}_1, \cdots, \boldsymbol{q}_{M-1}\right]$ represent the intermediate waypoints, $\overline{\boldsymbol{t}}=\left[\overline{t}_1, \cdots, \overline{t}_M\right]^\top$ are the time allocated for each piece, and $t_M$ is the total time. Given a set of $(\boldsymbol{Q}, \overline{\boldsymbol{t}})$, we can obtain a unique trajectory of minimum control effort in polynomial form through the mapping $\mathcal{C}(\cdot)$. This mapping is achieved by solving a boundary-intermediate value problem described in \cite{wang_geometrically_2022}, which returns the coefficients $\boldsymbol{C}=\left[\boldsymbol{C}_1^\top, \cdots, \boldsymbol{C}_M^\top \right]^\top$ with linear time and space complexity. Based on $(\boldsymbol{C}, \overline{\boldsymbol{t}})$, for a system of $S$ order integrator chain \cite{wang_geometrically_2022}, a polynomial trajectory $\boldsymbol{p}(t)$ of $N=2S-1$ degree can be defined as
\begin{align}
    \boldsymbol{p}(t)&=
    \begin{cases}
        \boldsymbol{p}_1\left(t-t_{0}\right), & \text{if } t \in\left[t_{0}, t_1\right) \\
        \cdots \\
        \boldsymbol{p}_i\left(t-t_{i-1}\right), & \text{if } t \in\left[t_{i-1}, t_i\right) \\
        \cdots \\
        \boldsymbol{p}_M\left(t-t_{M-1}\right), & \text{if } t \in\left[t_{M-1}, t_M\right) \\
    \end{cases},\\[5pt]
    \boldsymbol{p}_i(t)&=\boldsymbol{C}_i^\top \cdot \boldsymbol{\beta}(t), \quad \forall t \in\left[0, \overline{t}_i\right],
\end{align}
where $\boldsymbol{C}_i=\left[\boldsymbol{c}_{i,1},\cdots,\boldsymbol{c}_{i,D}\right] \in \mathbb{R}^{(N+1)\times D}$ is the coefficient matrix of the $i^{t h}$ piece, $\boldsymbol{\beta}(t):=\left[1, t, \cdots, t^N\right]^\top \in \mathbb{R}^{N+1}$ is the natural basis.

Based on the above parameterization, MINCO's objective $\mathcal{H}(\boldsymbol{Q}, \overline{\boldsymbol{t}})$ can be computed as 
\begin{equation}
\mathcal{H}(\boldsymbol{Q}, \overline{\boldsymbol{t}}) := K =\mathcal{K}(\mathcal{C}(\boldsymbol{Q}, \overline{\boldsymbol{t}}), \overline{\boldsymbol{t}}).
\end{equation}

For any second-order continuous cost function $\mathcal{K}(\boldsymbol{C}, \overline{\boldsymbol{t}})$, we can compute $\partial \mathcal{H} / \partial \boldsymbol{Q}$ and $\partial \mathcal{H} / \partial \overline{\boldsymbol{t}}$ from $\partial \mathcal{K} / \partial \boldsymbol{C}$ and $\partial \mathcal{K} / \partial \overline{\boldsymbol{t}}$ \cite{wang_geometrically_2022}, and use gradient descent to optimize the objective.

\subsection{Spatial-Temporal Optimization}
\label{optimization}

We construct the trajectory problem in the form of unconstrained optimization:
\begin{equation}
\min _{\boldsymbol{Q}, \overline{\boldsymbol{t}}} \mathcal{K}(\mathcal{C}(\boldsymbol{Q}, \overline{\boldsymbol{t}}), \overline{\boldsymbol{t}}):=\sum_x \omega_x K_x,
\label{object_function}
\end{equation}
where subscripts $x \in \{e, t, o, d \}$ stands for \textit{Control Effort} ($e$), \textit{Trajectory Time} ($t$), \textit{Obstacle Avoidance} ($o$), and \textit{Dynamical Feasibility} ($d$). $\omega_x$ are the weights for different costs.

The cost function $K_x$ and its gradients are composed of the cost and gradients of each trajectory piece:
\begin{gather}
    K_x=\sum_{i=1}^M K_x^i,\\
    \frac{\partial K_x}{\partial \boldsymbol{C}}=\left[\frac{\partial K_x^{1}}{\partial \boldsymbol{C}_1}^\top, \frac{\partial K_x^{2}}{\partial \boldsymbol{C}_2}^\top, \ldots, \frac{\partial K_x^{M}}{\partial \boldsymbol{C}_M}^\top\right]^\top,\\
    \frac{\partial K_x}{\partial \overline{\boldsymbol{t}}}=\left[\frac{\partial K_x^1}{\partial \overline{t}_1}, \frac{\partial K_x^2}{\partial \overline{t}_2}, \ldots, \frac{\partial K_x^M}{\partial \overline{t}_M}\right]^\top.
\end{gather}

The cost of control effort is calculated as the integral of the squared norm of the control input. The cost of trajectory time is determined by accumulating the trajectory time. For time-integral constraints, including obstacle avoidance and dynamical feasibility, we densely sample the trajectory, penalize violations of the constraints at each sample point, and accumulate the penalties to form the total cost. For a detailed formulation, we refer interested readers to \cite{wang_geometrically_2022}.

To solve the trajectory optimization problem (\ref{object_function}), we use the L-BFGS solver because it achieves rapid convergence. To avoid nonpositive values of $\overline{t}_i$, we introduce a proxy variable $\tau$ and modify $\overline{t}_i$ as
\begin{equation}
\overline{t}_i=\frac{\overline{t}_{\max }-\overline{t}_{\min }}{1+e^{-\tau_i}}+\overline{t}_{\min }, \quad i=1,2, \ldots, M.
\end{equation}
This establishes the mapping $\tau_i \in(-\infty,+\infty) \to \overline{t}_i \in\left(\overline{t}_{\min }, \overline{t}_{\max }\right)$, where $\overline{t}_{\max }$ and $\overline{t}_{\min }$ are the upper and lower bounds of the duration of the $i^{t h}$ piece trajectory that can be set according to the users' demand.

\subsection{Expert Planner and Multimodality}
\label{expert}
To provide training data for the NN-Planner, we implement an expert planner based on the optimization method \cite{quan_distributed_2022} presented in Section \ref{optimization}. During each replanning event, the expert planner receives the local initial state $\boldsymbol{S}_\text{init} = \left[ \boldsymbol{p}_\text{init}, \boldsymbol{v}_\text{init} \right] ^\top$ and the target state $\boldsymbol{S}_\text{target} = \left[ \boldsymbol{p}_\text{target}, \boldsymbol{v}_\text{target} \right] ^\top$ as inputs. To improve robustness in scenarios where local optima lie in different homotopy classes, the planner generates three initial trajectories—a straight line and two curves deformed toward the left and right—to capture the three fundamental motion patterns: straight, leftward, and rightward. Each trajectory is then optimized independently, and the planner selects the solution $\left( \boldsymbol{Q}^*, \overline{\boldsymbol{t}}^* \right)$ with the lowest cost among the three.

\begin{figure}[t]
    \setlength{\abovecaptionskip}{\aboveCaptionFig} 
    \setlength{\belowcaptionskip}{\belowCaptionFig}
    \centering
    \subfigure[Case 1]{
    \includegraphics[width=0.295\linewidth]{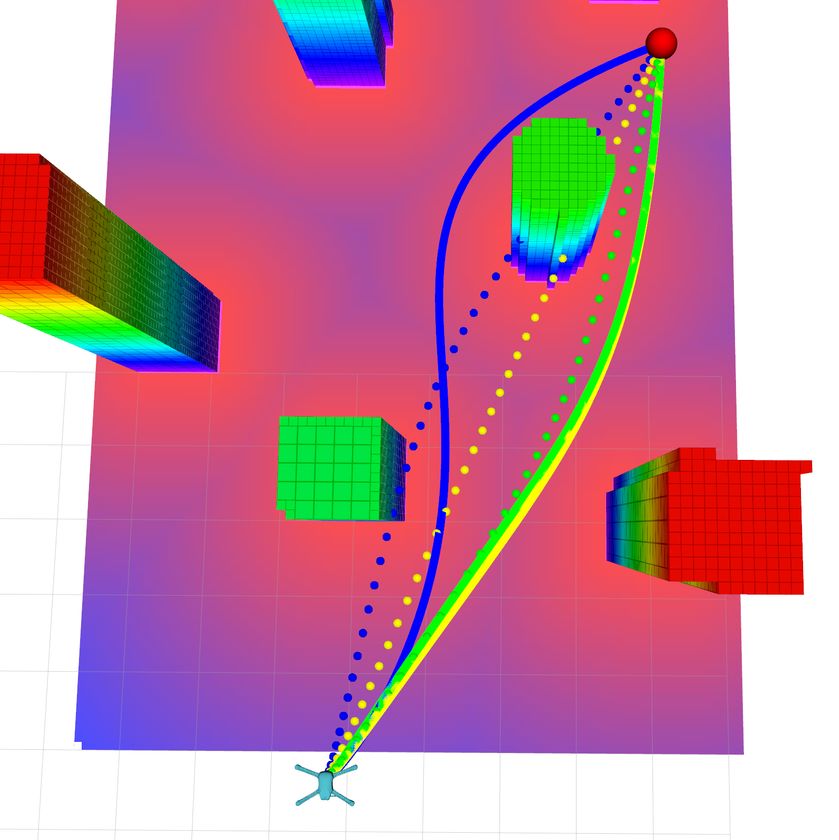}
    \label{case1}}
    \subfigure[Case 2]{
    \includegraphics[width=0.295\linewidth]{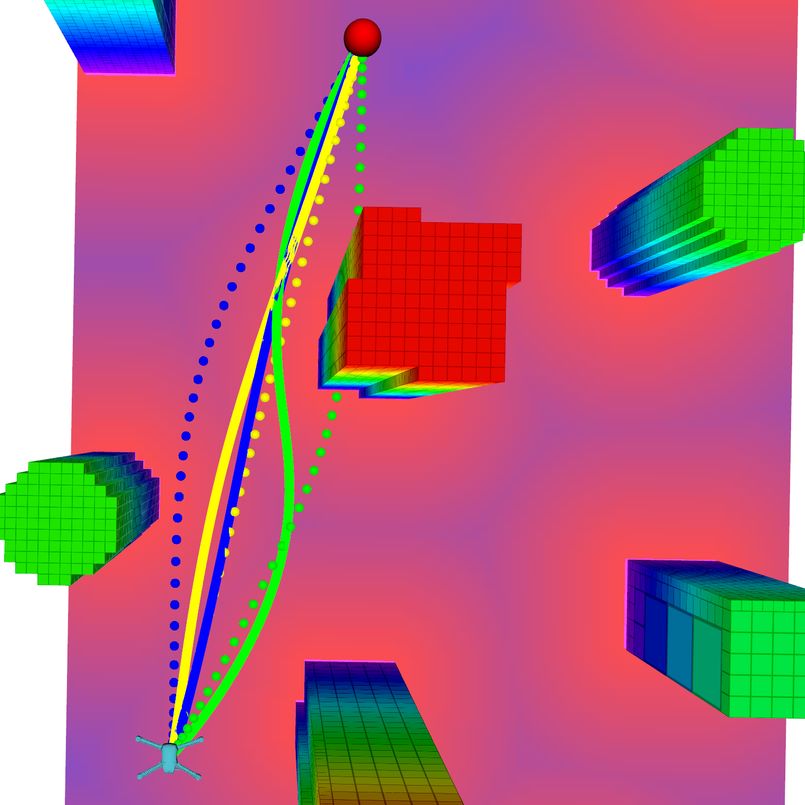}
    \label{case2}}
    \subfigure[Case 3]{
    \includegraphics[width=0.295\linewidth]{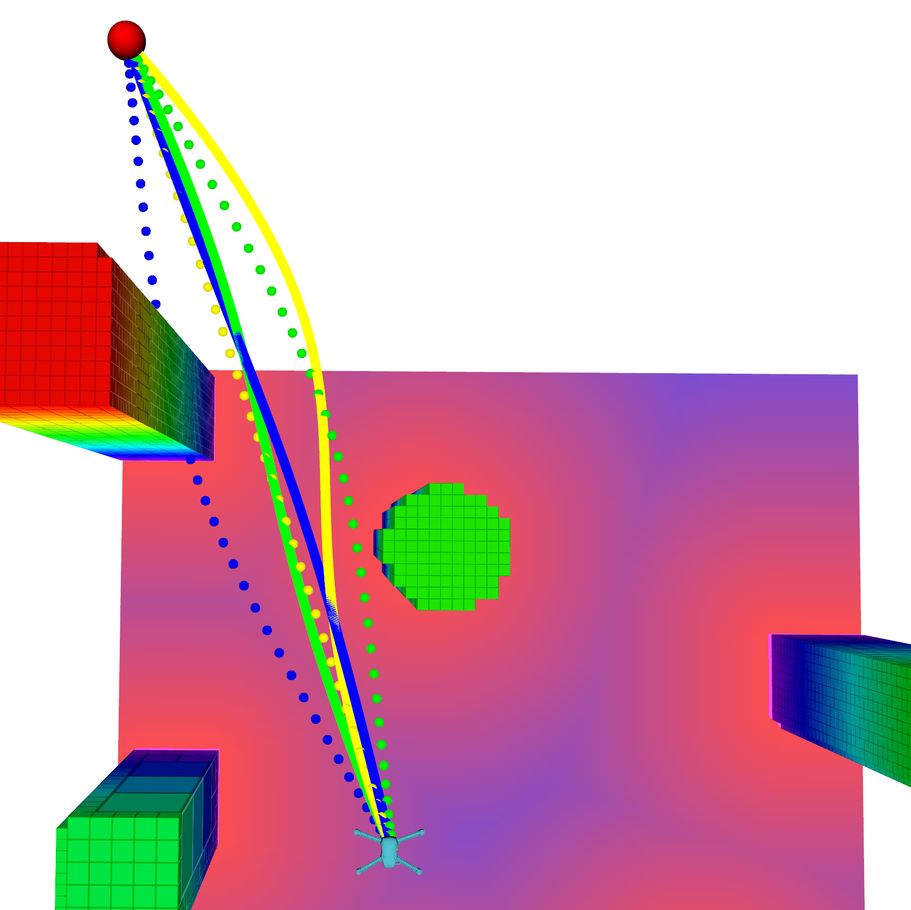}
    \label{case3}}
    \caption{\textbf{Illustration of the multimodality in three cases of local planning.} The optimization of the expert planner, starting from different initial trajectories (represented by dotted lines), may converge to distinct results (represented by solid lines).}
    \label{cases}
\end{figure}

The objective function of the optimization problem is not in an analytical form, introducing a nonconvex nature that significantly impacts both the solution process and the quality of the results. This sensitivity to the initial optimization value is illustrated in Fig. \ref{cases}, and its effect on both solution quality and computational cost is quantified in Table \ref{cost_of_cases}. These results highlight the necessity of incorporating a suitable initialization method.

\begin{table}[t]
\setlength{\abovecaptionskip}{\aboveCaptionFig} 
\setlength{\belowcaptionskip}{\belowCaptionFig}
\centering
\caption{Trajectory cost and computation time of optimization starting from different initial values}
\label{cost_of_cases}
\begin{tabular}{ccccc}
\toprule
\multirow{2}[3]{*}{\textbf{Case}}   & \multirow{2}[3]{*}{\textbf{Metric}} & \multicolumn{3}{c}{\textbf{Trajectory Color}} \\ \cmidrule{3-5} 
                        &                         & Yellow      & Green      & Blue      \\ 
\midrule
\multirow{2}{*}{Case 1} & Trajectory Cost         & \textbf{12.91}       & 12.93      & 14.75     \\
                        & Computation Time (s)    & 0.44        & \textbf{0.27}       & 0.49      \\ 
\midrule
\multirow{2}{*}{Case 2} & Trajectory Cost         & 10.87       & 26.62      & \textbf{10.83}     \\
                        & Computation Time (s)    & \textbf{0.23}        & 0.26       & 0.32      \\ 
\midrule
\multirow{2}{*}{Case 3} & Trajectory Cost         & 12.76       & 12.59      & \textbf{11.90}     \\
                        & Computation Time (s)    & \textbf{0.25}        & 0.40       & 0.36      \\
\bottomrule
\end{tabular}
\vspace{-9pt}
\end{table}

\subsection{Initialization: Neural Network Planner}
\label{nn_planner}
To mitigate the influence of the aforementioned nonconvexity on trajectory optimization, the NN-Planner's primary purpose is to capture the potential high-quality trajectories from raw sensory observations.
\subsubsection{\textbf{Structure}}
The NN-Planner takes in an observation
\begin{equation}
\boldsymbol{O} = \left(\boldsymbol{I}, { }^B\boldsymbol{v}_\text{drone}, { }_B^W \boldsymbol{R}, { }^B \boldsymbol{S}_\text{init}, { }^B \boldsymbol{S}_\text{target} \right),
\end{equation}
where $\boldsymbol{I}\in \mathbb{R}^{480 \times 640}$ is the depth image, ${ }^B \boldsymbol{v}_\text{drone} \in \mathbb{R}^3$ is the drone's velocity in body frame, ${ }_B^W \boldsymbol{R} \in \mathbb{R}^{3 \times 3}$ is the drone's attitude (rotation from body frame to world frame), ${ }^B \boldsymbol{S}_\text{init} \in \mathbb{R}^{2 \times 3}$ and ${ }^B \boldsymbol{S}_\text{target} \in \mathbb{R}^{2 \times 3}$ are the local initial state and target state in body frame, respectively. 

We have developed a neural network tailored for processing the given observation and producing the output $\left({}^B\hat{\boldsymbol{Q}}, {}^B\hat{\overline{\boldsymbol{t}}}\right)$. This observation encompasses both visual and inertial information and is processed through two distinct branches within the neural network, as illustrated in Fig. \ref{system_overview}. For the visual information $\boldsymbol{I}$, we utilize a pretrained ResNet-18 \cite{he_deep_2016} attached with a fully-connected layer to generate the visual feature in $\mathbb{R}^{24}$. 
In the case of the inertial information $\left( { }^B \boldsymbol{v}_\text{drone}, { }_B^W \boldsymbol{R}, { }^B \boldsymbol{S}_\text{init}, { }^B \boldsymbol{S}_\text{target} \right)$, we first flatten each of these elements and concanate them to form a $\mathbb{R}^{24}$ vector. This vector is then processed by a four-layer perceptron with $[48,24,24]$ hidden nodes to extract the inertial feature. Subsequently, the visual feature and inertial feature are concatenated and passed through another four-layer-perceptron with $[48,96,96]$ hidden nodes to generate the output vector. All multi-layer perceptrons employ the Leaky ReLU activation function. Finally, the estimated $\left( {}^B\hat{\boldsymbol{Q}}, {}^B\hat{\overline{\boldsymbol{t}}} \right)$ are derived from the output vector and transformed into the world frame. 

\subsubsection{\textbf{Data Acquisition and Training}}
We train the NN-Planner using supervised learning. We use the expert planner to collect training data in a self-built simulation environment, where each planning operation yields a training sample comprising $\boldsymbol{O}$ and the corresponding reference output $\left(^B\boldsymbol{Q}^*, ^B\overline{\boldsymbol{t}}^*\right)$ from the expert planner. We train the NN-Planner using Mean Squared Error (MSE) as the loss function and Adam \cite{kingma_adam_2015} as the optimizer. 

\subsection{Online Replanning Framework}
\label{online_framework}
\begin{figure}[bp]
    \vspace{-8pt}
    \setlength{\abovecaptionskip}{-3pt} 
    \setlength{\belowcaptionskip}{\belowCaptionFig}
    \centerline{\includegraphics[width=1\linewidth]{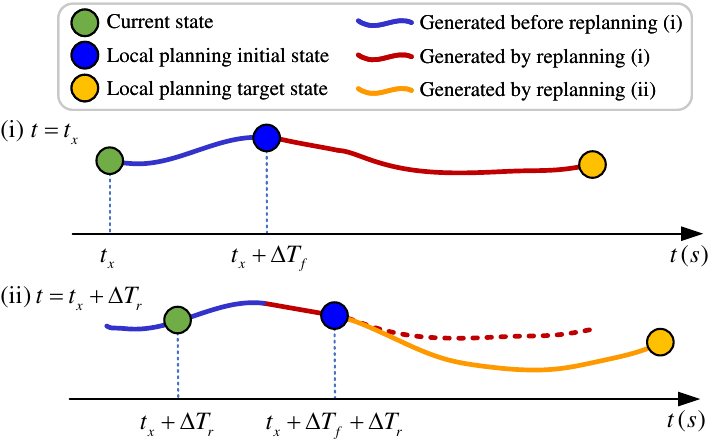}}
    \caption{Illustration of two consecutive replanning within the online replanning framework.}
    \label{replanning_framework}
\end{figure}

\begin{figure*}[t]
    \setlength{\abovecaptionskip}{\aboveCaptionFig} 
    \setlength{\belowcaptionskip}{\belowCaptionFig}
    \centering
    \subfigure[Scene 1: Poles]{
    \includegraphics[width=0.31\linewidth]{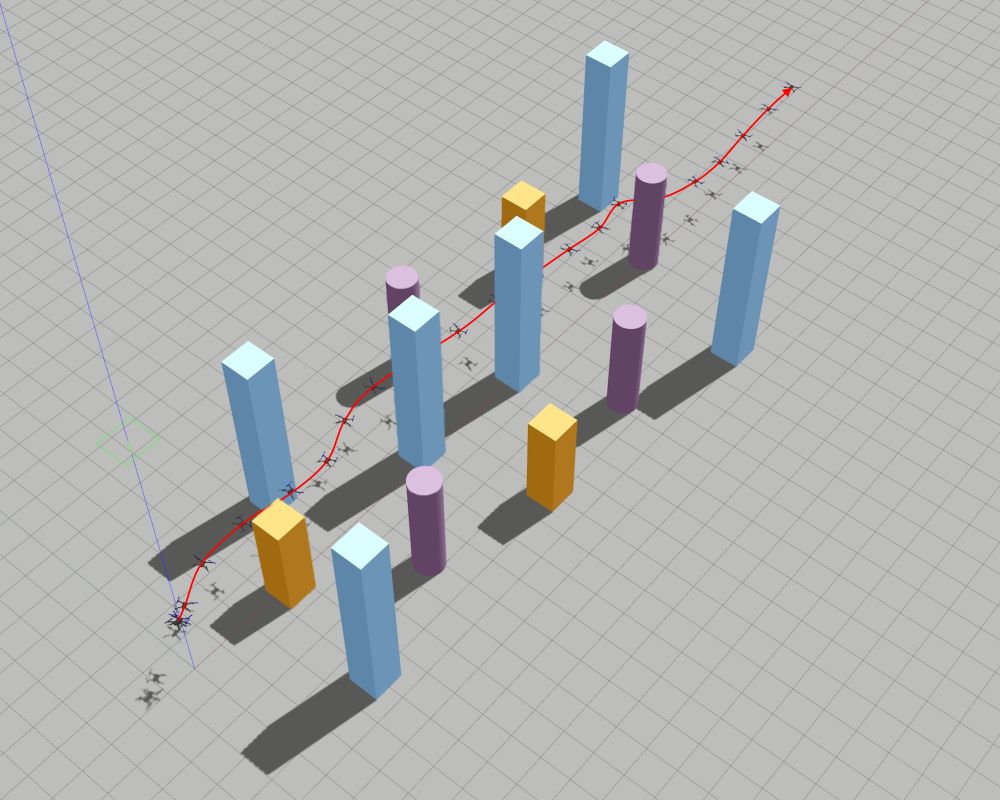}
    \label{poles}}
    \subfigure[Scene 2: Forest]{
    \includegraphics[width=0.31\linewidth]{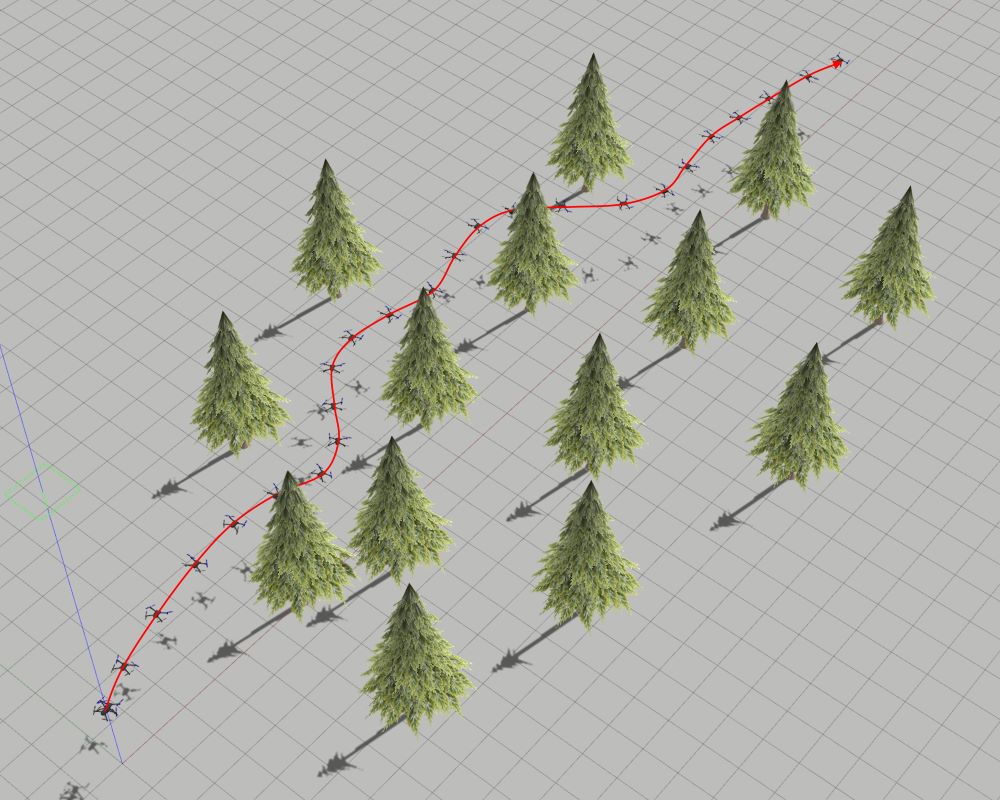}
    \label{forest}}
    \subfigure[Scene 3: Bricks]{
    \includegraphics[width=0.31\linewidth]{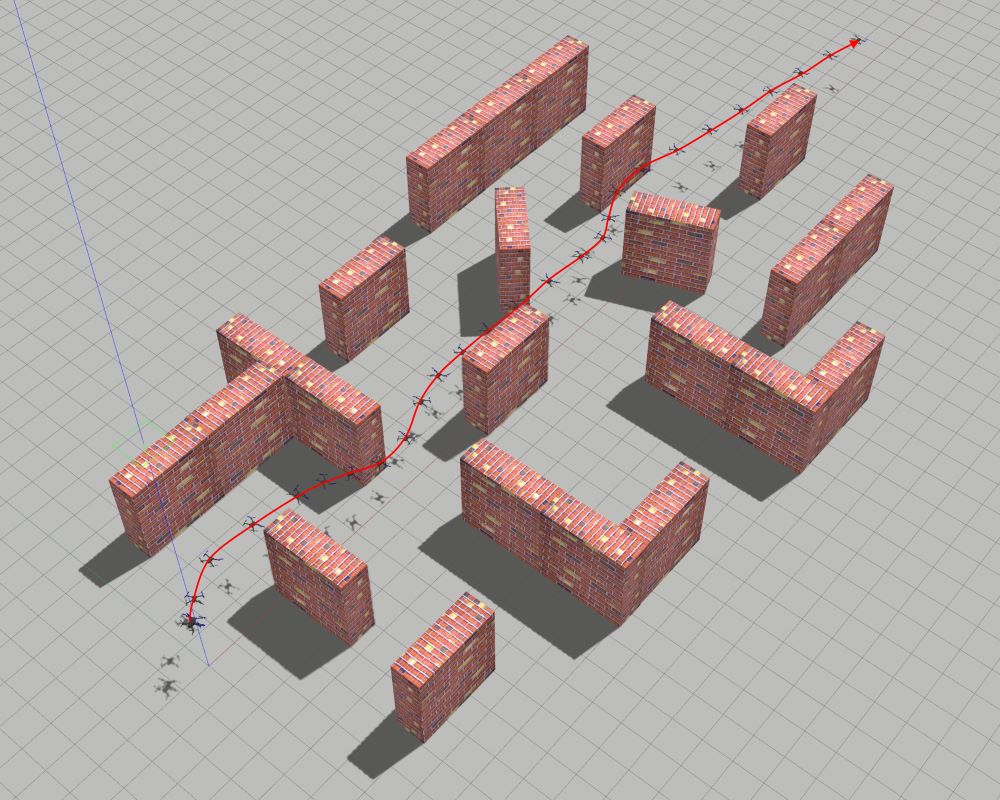}
    \label{bricks}}
    \subfigure[Scene 4]{
    \includegraphics[width=0.145\linewidth]{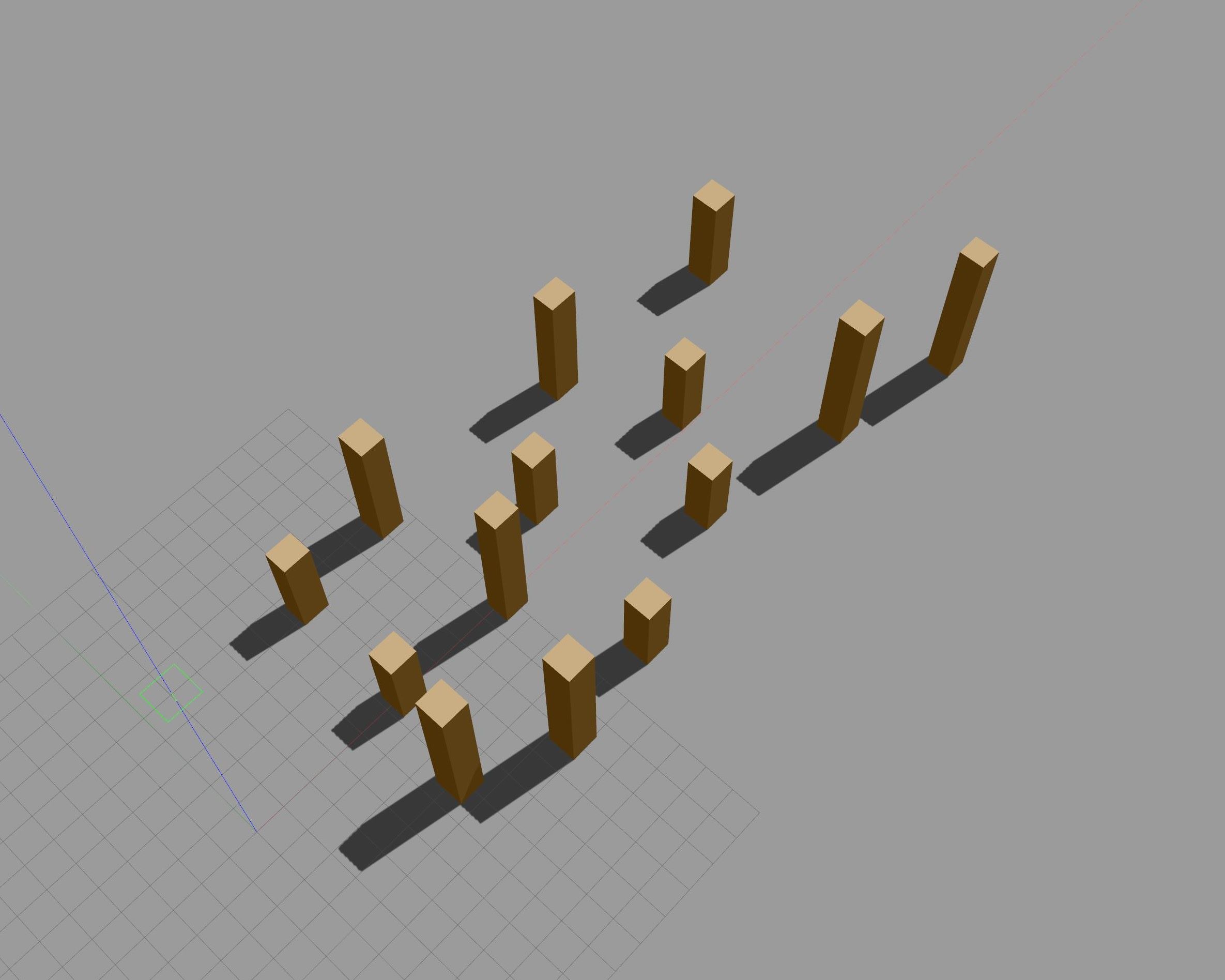}
    \label{scene4}}
    \subfigure[Scene 5]{
    \includegraphics[width=0.145\linewidth]{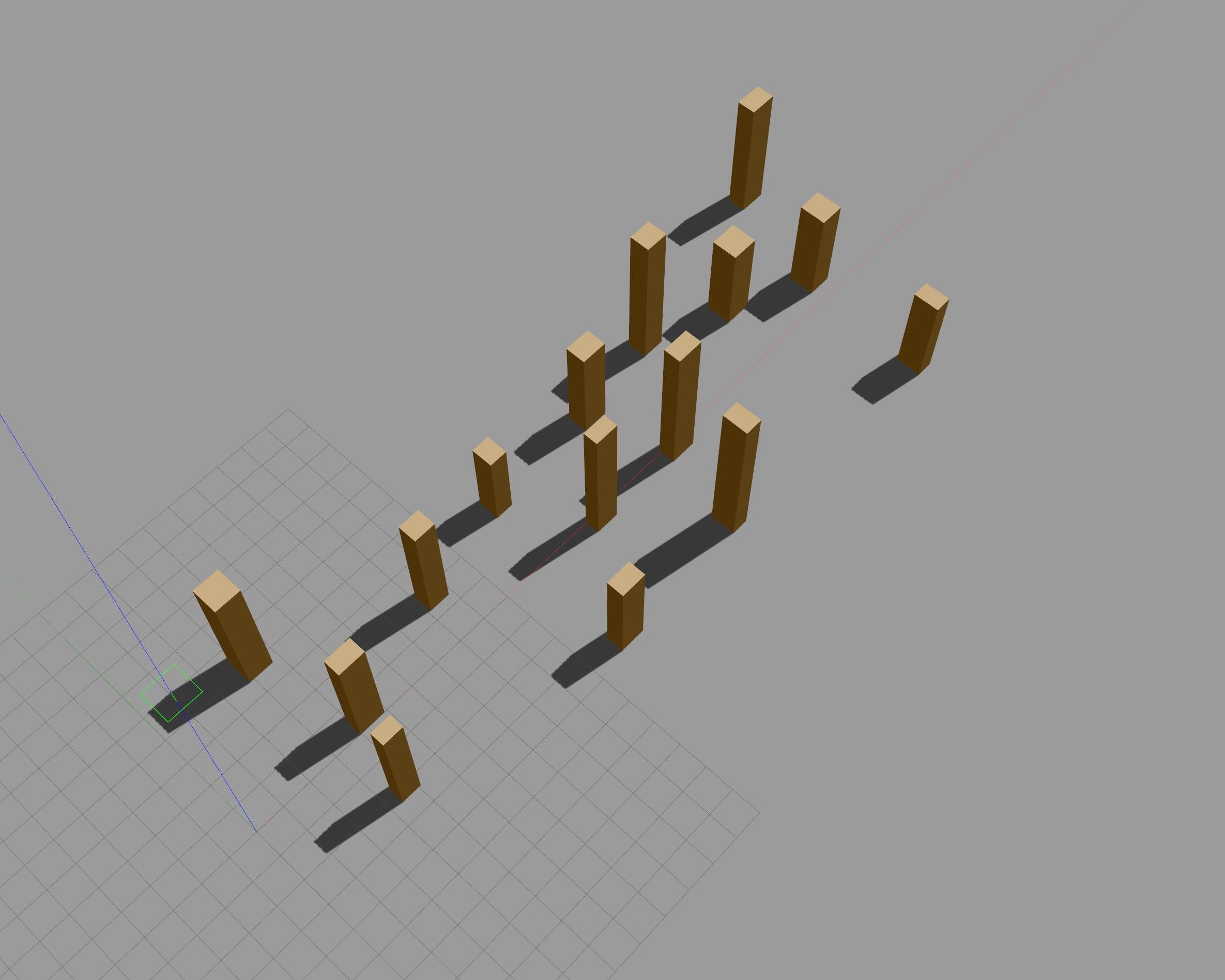}
    \label{scene5}}
    \subfigure[Scene 6]{
    \includegraphics[width=0.145\linewidth]{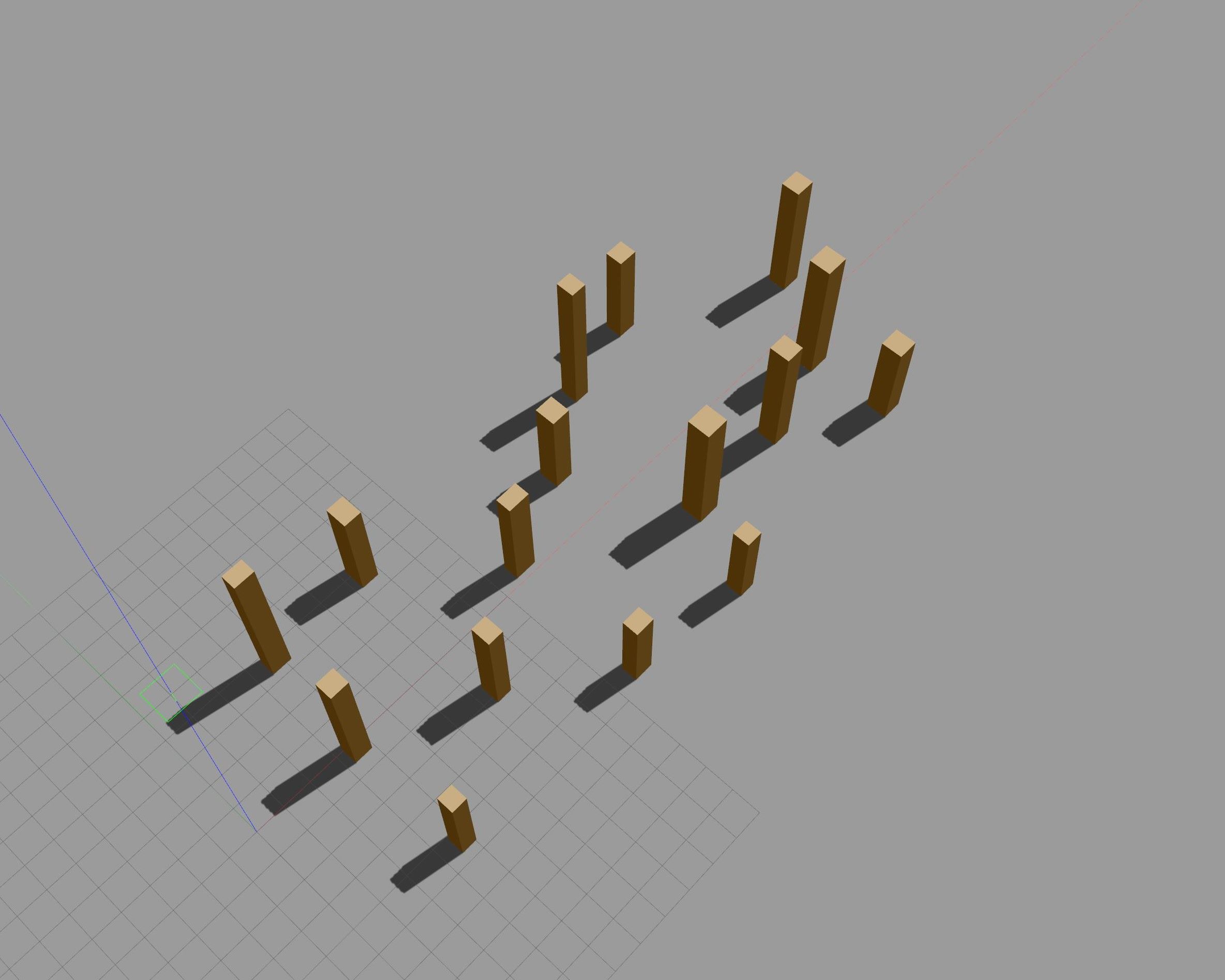}
    \label{scene6}}
    \subfigure[Scene 7]{
    \includegraphics[width=0.145\linewidth]{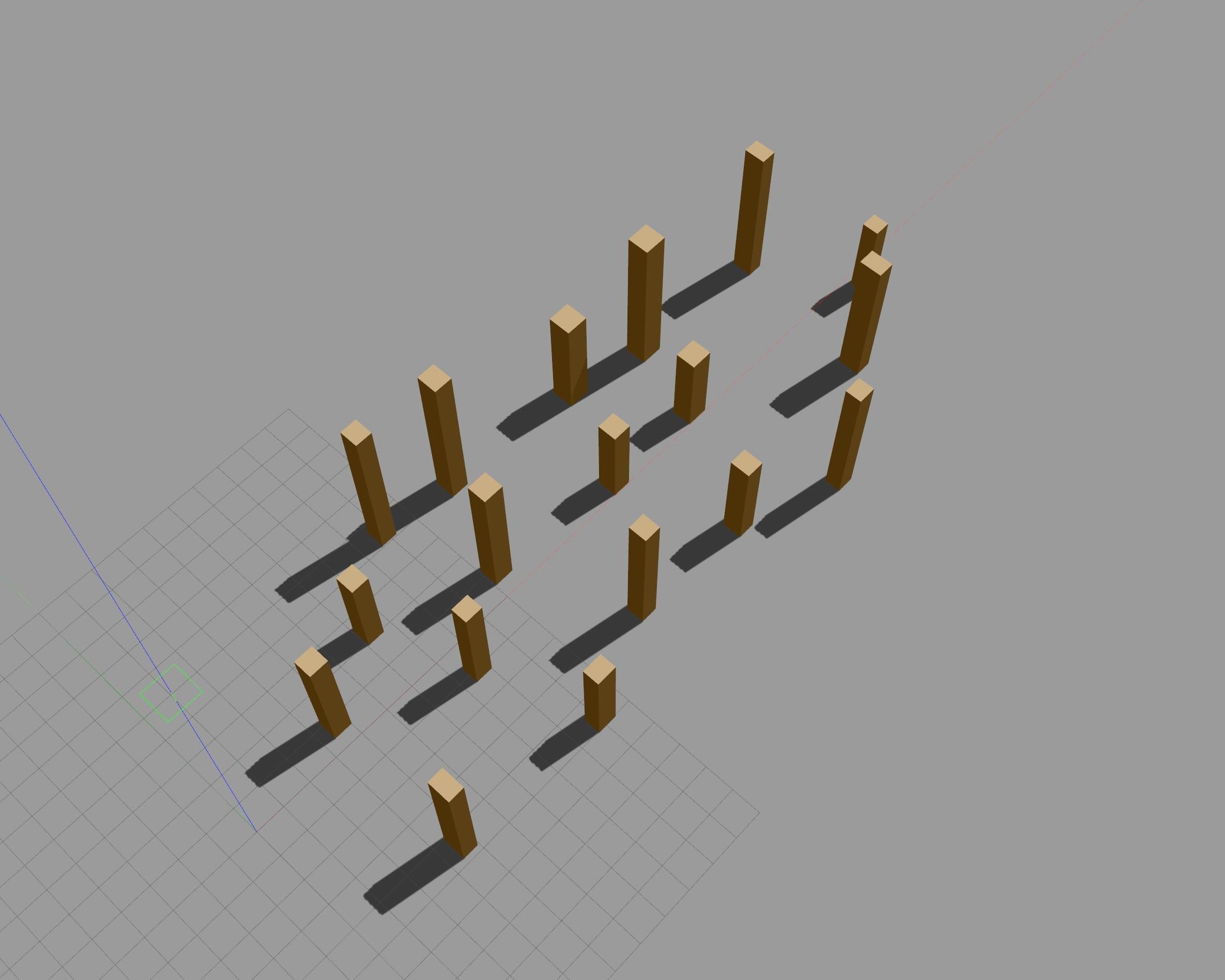}
    \label{scene7}}
    \subfigure[Scene 8]{
    \includegraphics[width=0.145\linewidth]{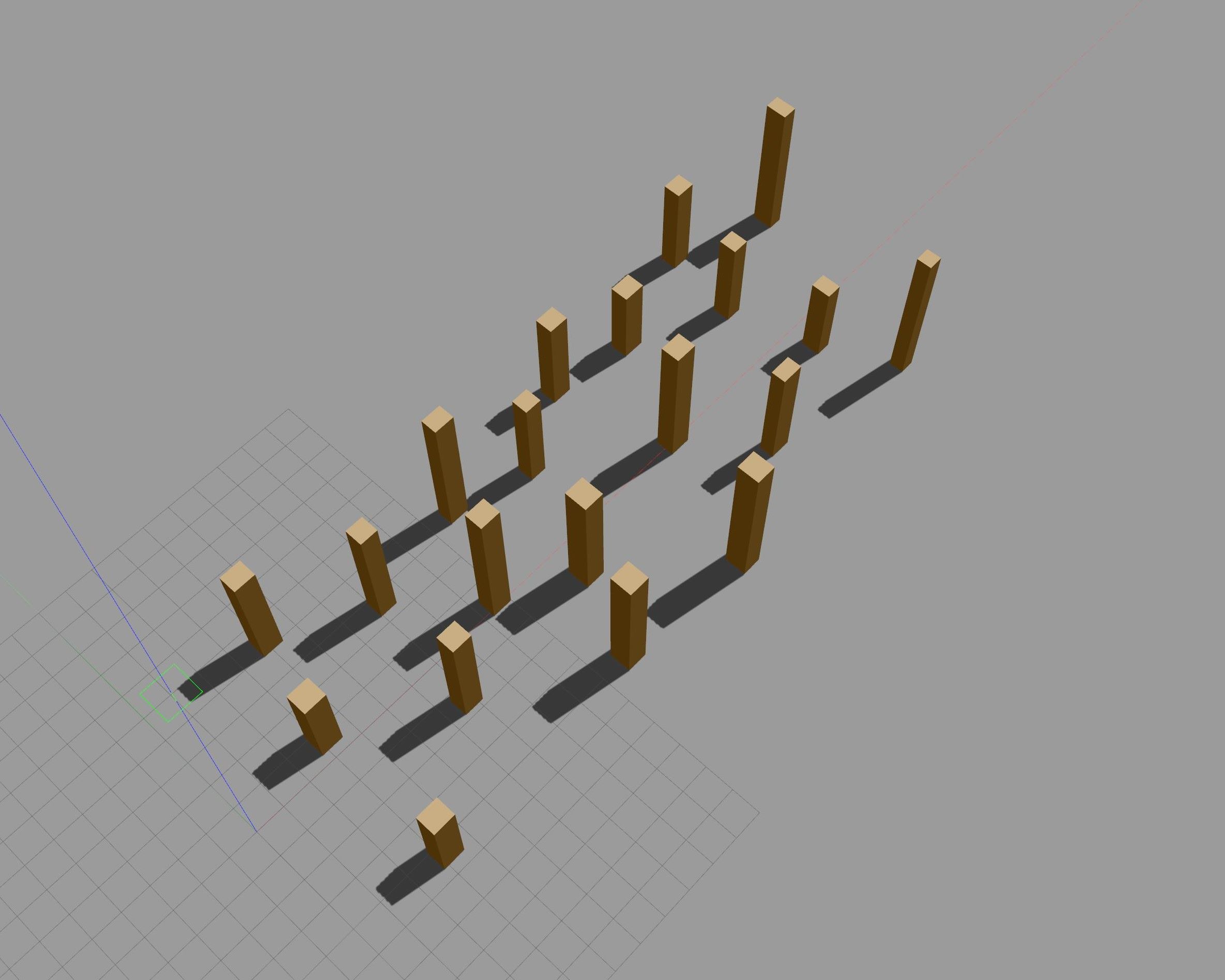}
    \label{scene8}}
    \subfigure[Scene 9]{
    \includegraphics[width=0.145\linewidth]{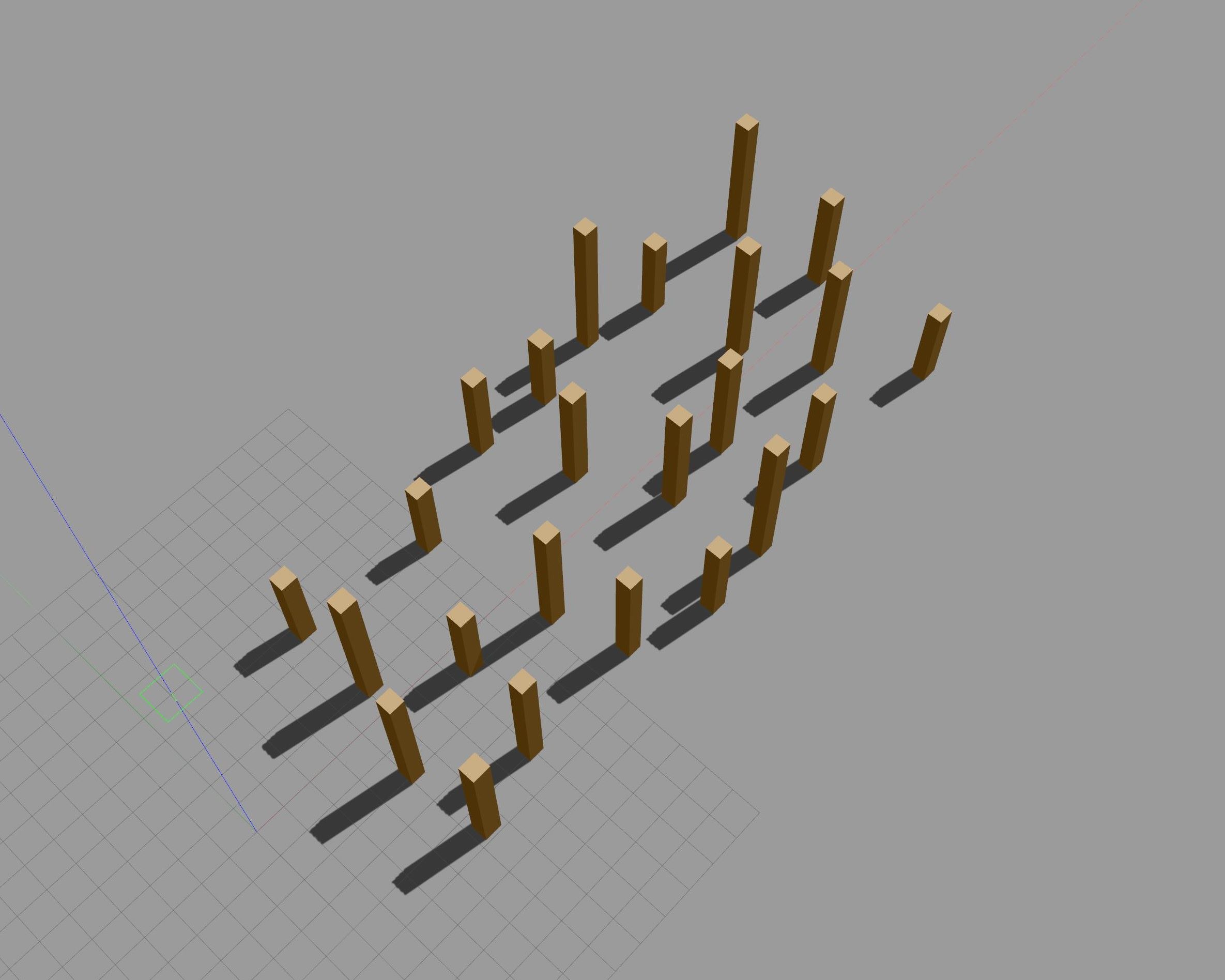}
    \label{scene9}}
    \caption{Nine scenes for simulation comparisons.}
    \vspace{-10pt}
    \label{scenes}
\end{figure*}

Autonomous flight in unknown environments requires online replanning because of the limited perception horizon. We design a robust online replanning framework that enables tolerance to planning latency. Within this framework, the planner maintains an evolving trajectory for the tracker to follow, as outlined in Fig. \ref{replanning_framework}.

In a replanning at time $t_x$, the planner first selects the local target state $\boldsymbol{S}_\text{target}$ and the local initial state $\boldsymbol{S}_\text{init}$. $\boldsymbol{S}_\text{target}$ comprises a collision-free point $\boldsymbol{p}_\text{init}$ at a specific distance ahead and a desired velocity $\boldsymbol{v}_\text{init}$. For the local initial state, the planner retrieves the state at $t_x+\Delta T_f$ along the existing trajectory as $\boldsymbol{S}_\text{init}$, where $\Delta T_f$ is the foreseeing horizon. The planner then generates a trajectory connecting them. This newly generated trajectory supersedes the portion of the existing trajectory beyond the time $t_x+\Delta T_f$. After an interval $\Delta T_r$ comes the next round of replanning, and the existing trajectory beyond the time $t_x+\Delta T_f+\Delta T_r$ is updated. The replanning interval $\Delta T_r$ can be a constant value, typically set as the upper bound of the estimated planning time, or a variable value based on real-time measurements recorded during each planning iteration.

This planning framework has two main advantages. First, it is robust to planning latency. The tracker continuously accesses the real-time trajectory state, while updates from the planner become available to the tracker after a foreseeing horizon $\Delta T_f$. As long as trajectory planning completes within this horizon, tracking remains seamless. Second, it decouples planning and control frequencies. The tracking controller can publish high-frequency control commands independently of the planner's update rate.

This approach relies on the assumption that each newly generated trajectory segment within the $\Delta T_f$ window is reliable. This assumption is reasonable as long as $\Delta T_f$ is not excessively long.

\section{Simulations}

To validate the effectiveness of the NN-Planner (Section \ref{nn_planner}) and the online replanning framework (Section \ref{online_framework}), we conduct a series of simulations.

\subsection{Simulation Scenario and Task Settings}

We conduct simulations in Gazebo across nine scenes. Scenes 1 to 3, named ``Poles", ``Forest", and ``Bricks", feature distinct obstacle types, while Scenes 4 to 9 are randomly generated with obstacles similar to those in ``Poles" but varying in size and density. In these random scenes, obstacles are positioned within a rectangular region bounded by (3, -5), (3, 5), (27, -5), and (27, 5) (unit: meters), with a minimum spacing of 1.8 meters and widths sampled from the range in Table \ref{obs_setting}. Fig. \ref{scenes} illustrates all nine scenes.

\begin{table}[b]
\setlength{\abovecaptionskip}{\aboveCaptionFig}
\setlength{\belowcaptionskip}{\belowCaptionFig}
\centering
\caption{Obstacle settings in six random scenes}
\label{obs_setting}
\begin{tabular}{ccc}
\toprule
\textbf{Scene ID} & \textbf{Obstacle Number} & \textbf{Obstacle Width Range (m)} \\ \midrule
4                 & 14                & $[0.8, 1.0]$           \\
5                 & 15                & $[0.5, 1.0]$                \\
6                 & 16                & $[0.5, 0.8]$                \\
7                 & 18                & $[0.5, 0.8]$                \\
8                 & 20                & $[0.5, 0.8]$                \\
9                 & 24                & $[0.5, 0.6]$                \\ \bottomrule
\end{tabular}
\end{table}

In each simulation, the drone autonomously flies at a fixed altitude of 2 meters from a start point at (0, 0) to a target point at (30, 0).

\subsection{The Effectiveness of NN-Planner}

\subsubsection{\textbf{Implementation Details}}
To validate the neural network's ability to accelerate computation and generalize to unseen environments, we perform simulations using PX4 software-in-the-loop (SITL). The drone navigates using only onboard observations, without prior environmental knowledge. We compare three algorithms that share the same optimization framework (Section \ref{optimization}), differing only in how they generate initial trajectories during replanning:

a) Baseline Planner \cite{wang_geometrically_2022}: The initial waypoints $\boldsymbol{Q}$ are uniformly distributed between the start and target positions. The initial time allocation $\overline{\boldsymbol{t}}$ assigns equal durations to all trajectory segments, except the first and last, which are 1.5 times longer.

b) Geo-Planner: The initial waypoints $\boldsymbol{Q}$ are generated by A* algorithm, while the time allocation $\overline{\boldsymbol{t}}$ follows the same initialization as the Baseline Planner.

c) NEO-Planner (Proposed): Both spatial and temporal profiles are initialized by the NN-Planner. The neural network is trained exclusively on data from Scene 1 (``Poles") and deployed using ONNX-Runtime for fast inference.

Table \ref{params} summarizes the optimization parameters. The trajectory tracking controller derives position, velocity, and acceleration commands from the planned trajectory. For perception-aware navigation, it aligns the desired yaw angle tangentially to the trajectory. All commands are published to the PX4 controller at 60 Hz.

\subsubsection{\textbf{Metrics}}
We evaluate performance using four metrics, reflecting reliability, trajectory quality, computational cost, and the quality of the initial trajectory estimate:

a) Success Rate: the ratio of successful runs to total runs in repeated simulations. A run is successful if the drone reaches the target without colliding with obstacles.

b) Trajectory Cost: computed by recording the drone's state every 0.5 seconds. The cost includes the trajectory length, penalties for obstacle collisions, and violations of dynamical feasibility, all with equal weights.

c) Average Replanning Time: the mean replanning computation time per run, accounting for multiple replanning events within each run.

d) Average Optimization Iterations: the mean number of L-BFGS solver iterations per replanning event to solve the optimization problem (\ref{object_function}). With all other solver factors—including termination criteria—held constant, a lower iteration count implies a better initial guess.

\begin{table}[t]
\vspace{6pt}
\setlength{\abovecaptionskip}{\aboveCaptionFig} 
\setlength{\belowcaptionskip}{\belowCaptionFig}
\centering
\caption{Main parameters of the planners}
\label{params}

\begin{tabular}{%
    cc@{\hspace{0.5em}\vrule width 0.8pt\hspace{0.5em}}%
    cc%
  }
\toprule
\textbf{Parameter}  & \textbf{Value}
  & \textbf{Parameter}     & \textbf{Value}       \\
\midrule
$M$        & 3
  & $\overline{t}_{\min}$ & 0.5\,\text{s}   \\
$D$        & 2
  & $\overline{t}_{\max}$ & 5.0\,\text{s}   \\
$S$        & 3
  & $\Delta T_r$          & 1.0\,\text{s}   \\
$v_{\max}$ & 1.0\,\text{m/s}
  & $\Delta T_f$          & 1.0\,\text{s}   \\
\multicolumn{2}{l}{Weights between costs in (\ref{object_function})}
  & \multicolumn{2}{c}{$[1,1,10000,1]$}   \\
\bottomrule
\end{tabular}
\vspace{-19pt}
\end{table}

\begin{figure}[b]
\vspace{-6pt}
\setlength{\abovecaptionskip}{\aboveCaptionFig} 
\setlength{\belowcaptionskip}{\belowCaptionFig}
\centerline{\includegraphics[width=1\linewidth]{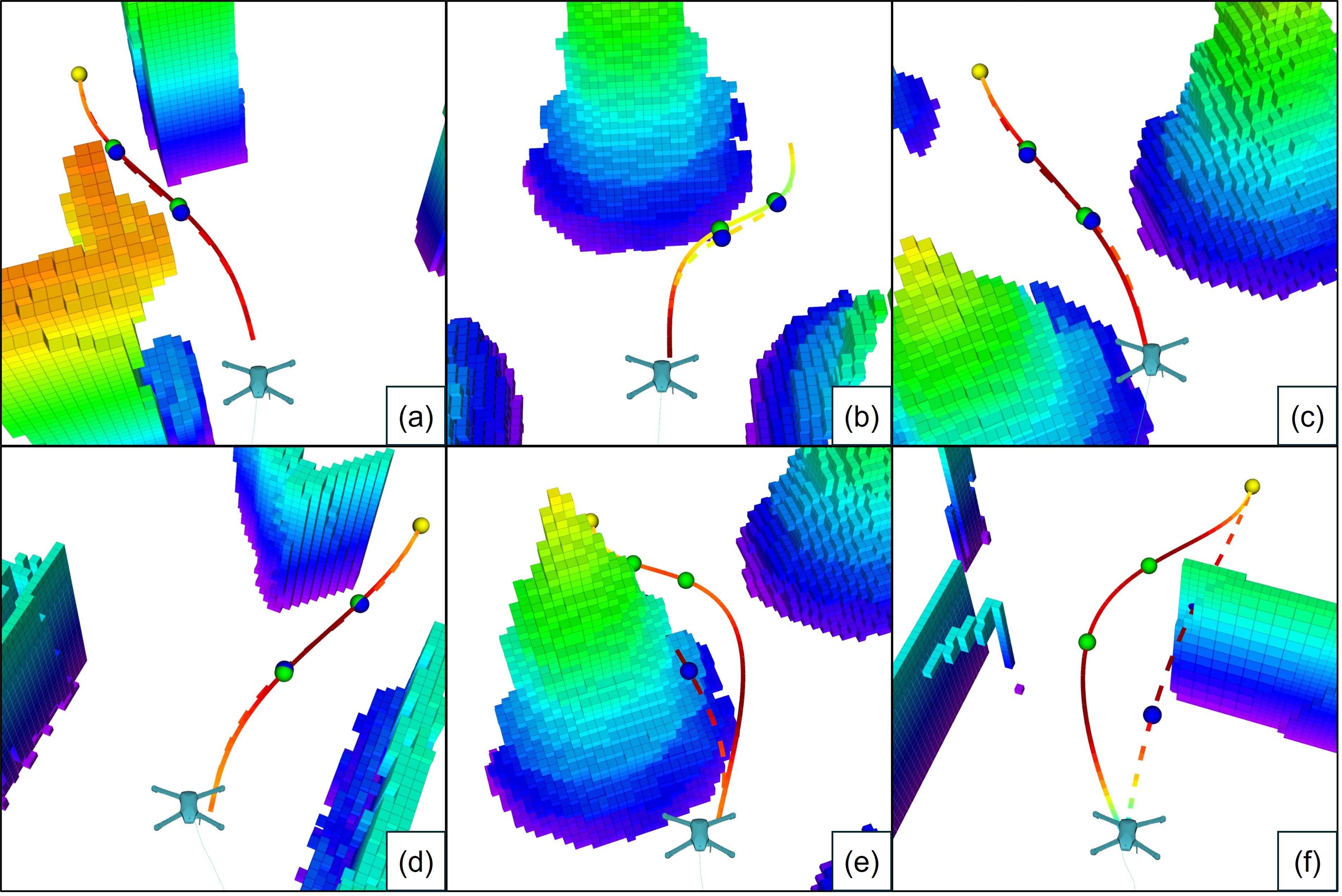}}
\caption{\textbf{Examples of neural network predicted trajectories and corresponding optimized results.} The dashed lines with blue spherical markers represent neural network predictions, while solid lines with green spherical markers denote optimized trajectories. The network usually outputs near-optimal trajectories, with any infeasibility corrected by optimization to maintain safe flight.}
\label{nn_prediction}
\end{figure}

\subsubsection{\textbf{Results and Discussions}}

Across all nine scenarios, we conduct 20 independent simulations per trajectory planning algorithm, totaling 540 runs. The results, summarized in Table \ref{planning_results}, show that the proposed planner achieves the same success rate as the Baseline Planner while maintaining a trajectory cost only 1\% higher. Furthermore, it reduces optimization iterations by 20\%, leading to a 26\% decrease in replanning time.

\begin{table}[t]
\vspace{6pt}
\setlength{\abovecaptionskip}{\aboveCaptionFig} 
\setlength{\belowcaptionskip}{\belowCaptionFig}
\centering
\caption{Planning results of three algorithms}
\label{planning_results}
\begin{tabular}{ccccc}
\toprule
\multirow{2}{*}{\textbf{Scene ID}} & \multirow{2}{*}{\textbf{Metric}}         & \multicolumn{3}{c}{\textbf{Planning Algorithms}}      \\ \cmidrule{3-5}
                          &                                 & Baseline & Geo & \textbf{NEO} \\ \midrule
\multirow{4}{*}{1}        & Success Rate                    & \textbf{0.9}              & 0.85        & \textbf{0.9}         \\
                          & Avg. Trajectory Cost         & 39.1             & 37.5        & \textbf{36.8}        \\
                          & Avg. Replanning Time (s)     & 0.179            & 0.151       & \textbf{0.119}       \\
                          & Avg. Opt. Iterations & 10.6             & 11.6        & \textbf{8.46}        \\ \midrule
\multirow{4}{*}{2}        & Success Rate                    & 0.75             & 0.65        & \textbf{0.8}         \\
                          & Avg. Trajectory Cost         & \textbf{35.9}             & 36.3        & 37.2        \\
                          & Avg. Replanning Time (s)     & 0.169            & 0.168       & \textbf{0.132}       \\
                          & Avg. Opt. Iterations & 10.1             & 10          & \textbf{8.25}        \\ \midrule
\multirow{4}{*}{3}        & Success Rate                    & 0.6              & \textbf{0.7}         & 0.65        \\
                          & Avg. Trajectory Cost         & \textbf{35.6}             & 37.5        & 37.1        \\
                          & Avg. Replanning Time (s)     & 0.201            & 0.192       & \textbf{0.143}       \\
                          & Avg. Opt. Iterations & 9.81             & 9.88        & \textbf{8.12}        \\ \midrule
\multirow{4}{*}{4}        & Success Rate                    & 0.95             & \textbf{1.0}           & 0.95        \\
                          & Avg. Trajectory Cost         & 38.2             & \textbf{36.3}        & 39          \\
                          & Avg. Replanning Time (s)     & 0.137            & 0.134       & \textbf{0.0923}      \\
                          & Avg. Opt. Iterations & 11.3             & 12.6        & \textbf{9.01}        \\ \midrule
\multirow{4}{*}{5}        & Success Rate                    & \textbf{1.0}                & 0.95        & 0.9         \\
                          & Avg. Trajectory Cost         & 38.7             & \textbf{37.8}        & 39.1        \\
                          & Avg. Replanning Time (s)     & 0.119            & 0.136       & \textbf{0.0893}      \\
                          & Avg. Opt. Iterations & 11.5             & 12.6        & \textbf{8.86}        \\ \midrule
\multirow{4}{*}{6}        & Success Rate                    & 0.9              & \textbf{1.0}           & 0.85        \\
                          & Avg. Trajectory Cost         & \textbf{37.9}             & 39          & 38.2        \\
                          & Avg. Replanning Time (s)     & 0.157            & 0.142       & \textbf{0.117}       \\
                          & Avg. Opt. Iterations & 11.5             & 11.4        & \textbf{9.29}        \\ \midrule
\multirow{4}{*}{7}        & Success Rate                    & 0.65             & 0.7         & \textbf{0.85}        \\
                          & Avg. Trajectory Cost         & 37.8             & 39.6        & \textbf{37.4}        \\
                          & Avg. Replanning Time (s)     & 0.191            & 0.18        & \textbf{0.144}       \\
                          & Avg. Opt. Iterations & 10.2             & 9.94        & \textbf{8.04}        \\ \midrule
\multirow{4}{*}{8}        & Success Rate                    & \textbf{0.95}             & 0.9         & 0.8         \\
                          & Avg. Trajectory Cost         & 38.1             & \textbf{37.8}        & 39          \\
                          & Avg. Replanning Time (s)     & 0.202            & 0.184       & \textbf{0.171}       \\
                          & Avg. Opt. Iterations & 10.2             & 10.2        & \textbf{8.4}         \\ \midrule
\multirow{4}{*}{9}        & Success Rate                    & \textbf{0.95}             & 0.75        & \textbf{0.95}        \\
                          & Avg. Trajectory Cost         & 38.2             & \textbf{37.3}        & 39          \\
                          & Avg. Replanning Time (s)     & 0.163            & 0.17        & \textbf{0.12}        \\
                          & Avg. Opt. Iterations & 9.81             & 9.38        & \textbf{7.55}        \\
\bottomrule
\end{tabular}
\vspace{-13pt}
\end{table}

NEO-Planner demonstrates the shortest average planning time across all nine simulation scenarios while maintaining a comparable success rate and trajectory cost to the other algorithms. This confirms that NEO-Planner improves computational efficiency without compromising reliability or trajectory quality. Additionally, NEO-Planner consistently requires the fewest optimization iterations, explaining its computational speed advantage: the neural network provides well-initialized trajectories, reducing the need for extensive optimization. Fig. \ref{nn_prediction} illustrates this effect. In most cases—such as the first four examples—the neural network predicts a trajectory that closely matches the final optimized result. In a few instances, shown in the last two examples, we also observe that the network fails to produce a fully feasible trajectory. It nevertheless captures the overall trend. The subsequent optimization step then ensures flight safety.

In contrast, Geo-Planner uses A* for spatial initialization, but lacks effective time parameter prediction, limiting its optimization efficiency. By generating both spatial ($\boldsymbol{Q}$) and temporal ($\overline{\boldsymbol{t}}$) initial values, NEO-Planner enables trajectory optimization to start from a more favorable state, requiring fewer iterations and thus reducing computational time.

Regarding generalization, NEO-Planner was trained solely on Scene 1 (``Poles"), yet it performs consistently across Scenes 2–9, which feature varying obstacle sizes, densities, and visual characteristics. Despite these variations, Table \ref{planning_results} shows no significant decline in NEO-Planner's reliability or performance, demonstrating its generalization capability.

This robustness comes from two factors. First, NEO-Planner uses depth maps, which are insensitive to color and emphasize object shape, aiding generalization. Second, its optimization step refines the network’s output, correcting errors and boosting performance in unfamiliar scenes.

\subsection{Tolerance to Planning Latency}

\subsubsection{\textbf{Simulation Settings and Metrics}}

To validate the latency tolerance of the online replanning framework (Section \ref{online_framework}), we inject an extra 0.8 s planning delay and run ten trials in Scenes 1–3 both with and without ($\Delta T_f=0\,\text{s}$) the foreseeing horizon. In each trial, we sample the desired and actual position and velocity every 0.1 s and compute the Root Mean Square Error (RMSE) in tracking.

\subsubsection{\textbf{Results and Discussions}}

\begin{table}[t]
\vspace{6pt}
\setlength{\abovecaptionskip}{\aboveCaptionFig} 
\setlength{\belowcaptionskip}{\belowCaptionFig}
\centering
\caption{Tracking error comparisons with and without the foreseeing horizon}
\label{tracking_err}
\begin{tabular}{cccc}
\toprule
\multirow{2}[3]{*}{\textbf{Scene}}  & \multirow{2}[3]{*}{\textbf{Metric}} & \multicolumn{2}{c}{\textbf{Average RMSE}}        \\ \cmidrule{3-4} 
                        &                         & $\Delta T_f=0\,\text{s}$ & $\Delta T_f=1\,\text{s}$ \\ 
\midrule
\multirow{2}{*}{Poles}  & Position Error ($\text{m}$)    & 0.23           & \textbf{0.13}           \\
                        & Velocity Error ($\text{m/s}$)  & 0.21           & \textbf{0.08}           \\ 
\midrule
\multirow{2}{*}{Forest} & Position Error ($\text{m}$)    & 0.28           & \textbf{0.16}           \\
                        & Velocity Error ($\text{m/s}$)  & 0.26           & \textbf{0.12}           \\ 
\midrule
\multirow{2}{*}{Bricks} & Position Error ($\text{m}$)    & 0.21           & \textbf{0.10}           \\
                        & Velocity Error ($\text{m/s}$)  & 0.19           & \textbf{0.07}           \\ 
\bottomrule
\vspace{-21pt}
\end{tabular}
\end{table}

\begin{figure}[b]
    \vspace{-15pt}
    \setlength{\abovecaptionskip}{\aboveCaptionFig} 
    \setlength{\belowcaptionskip}{\belowCaptionFig}
    \centering
    \subfigure[$t=2\,\text{s}$]{
    \includegraphics[width=0.295\linewidth]{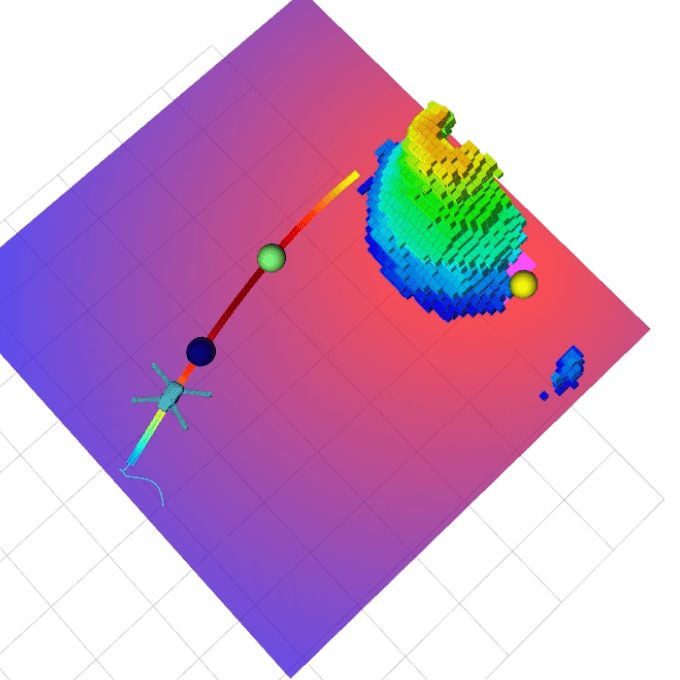}
    \label{tf0-t=2}}
    \subfigure[$t=4\,\text{s}$]{
    \includegraphics[width=0.295\linewidth]{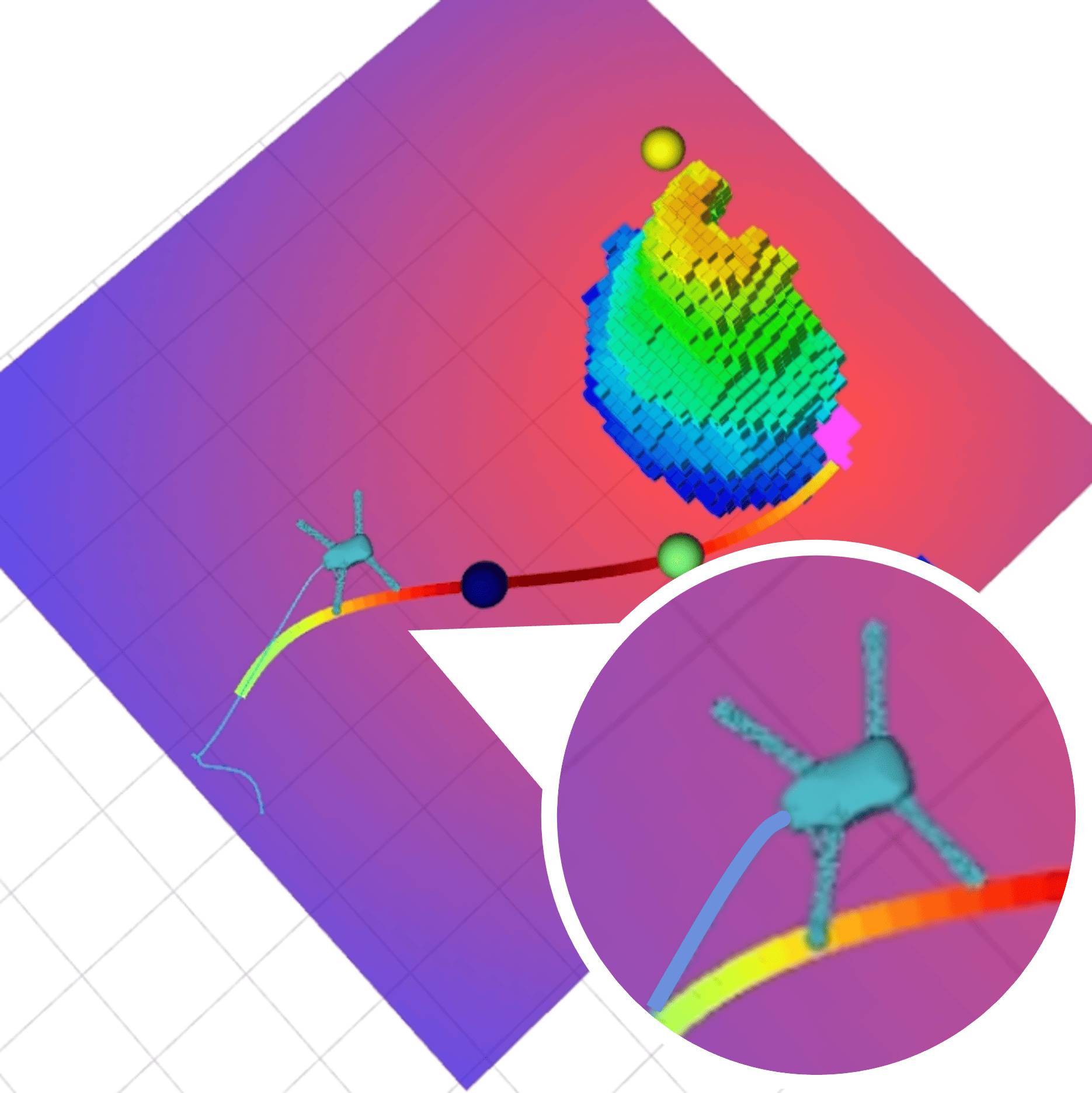}
    \label{tf0-t=4}}
    \subfigure[$t=5\,\text{s}$]{
    \includegraphics[width=0.295\linewidth]{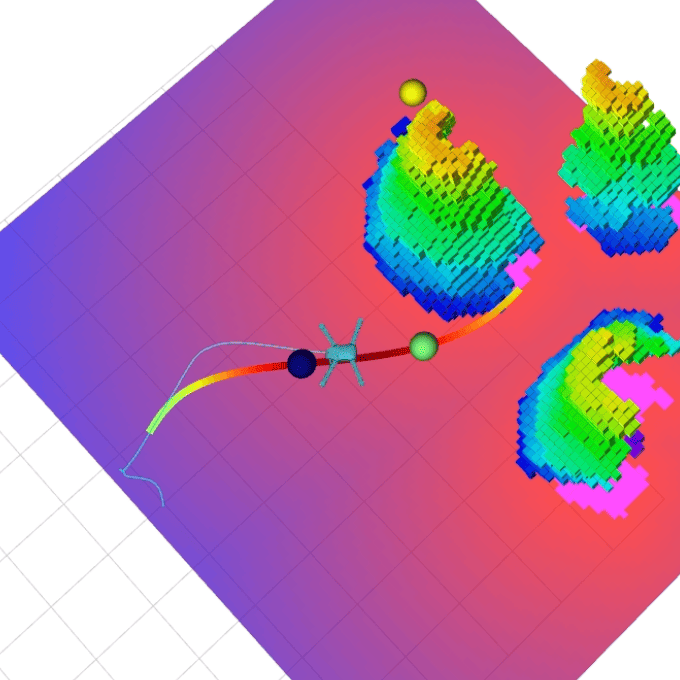}
    \label{tf0-t=5}}
    \subfigure[$t=6\,\text{s}$]{
    \includegraphics[width=0.295\linewidth]{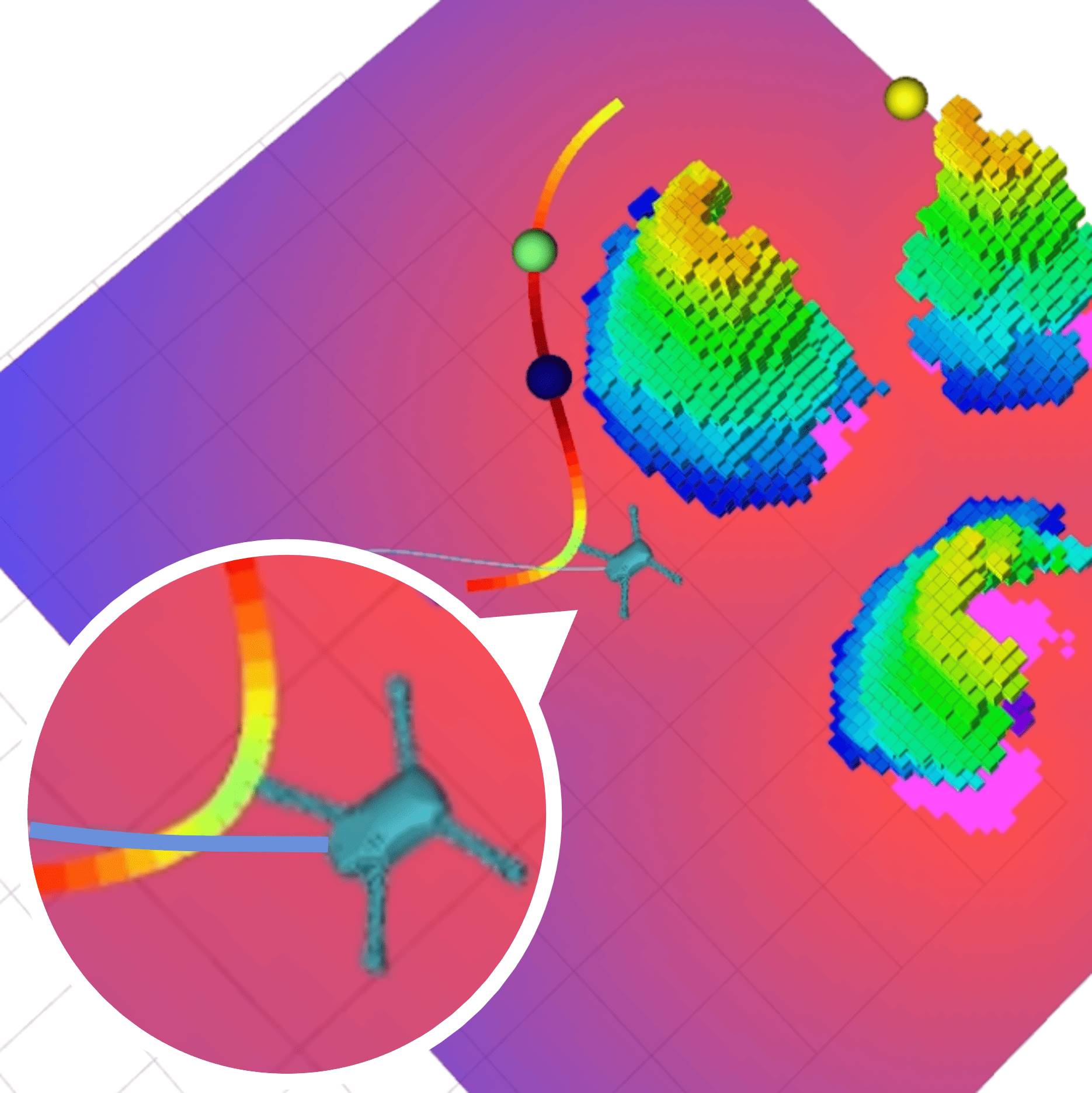}
    \label{tf0-t=6}}
    \subfigure[$t=7\,\text{s}$]{
    \includegraphics[width=0.295\linewidth]{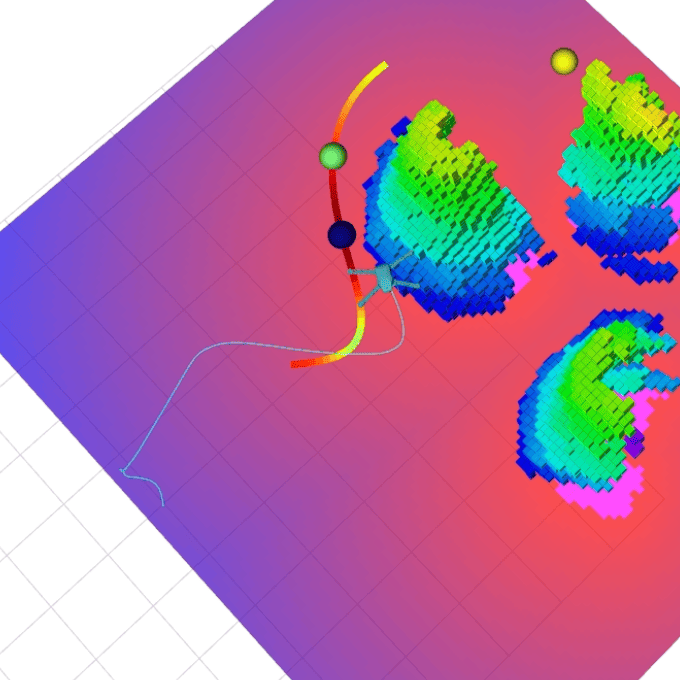}
    \label{tf0-t=7}}
    \subfigure[$t=8\,\text{s}$]{
    \includegraphics[width=0.295\linewidth]{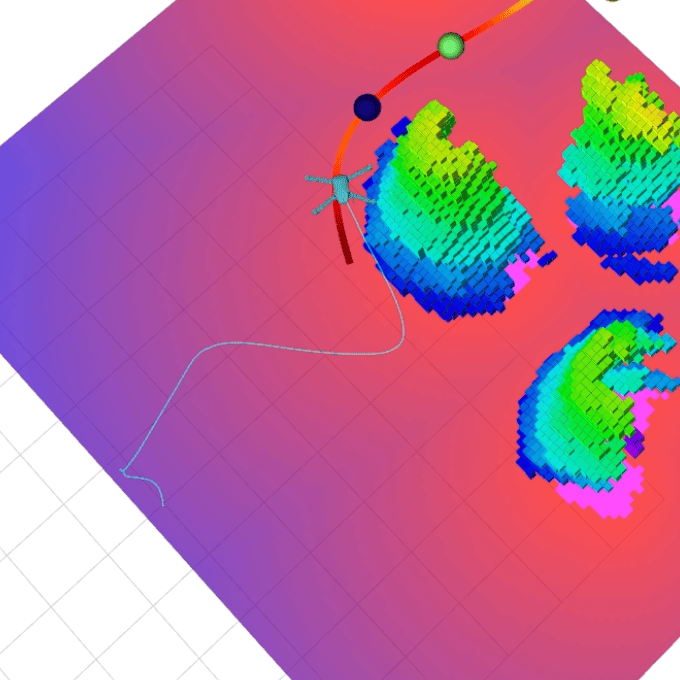}
    \label{tf0-t=8}}
    \caption{\textbf{Snapshots of one flight simulation in Scene 2 (Forest) with $\Delta T_f=0\,\text{s}$.} The thin light blue line represents the drone's actual flight trajectory, while the thick colored line with spherical markers denotes the continuously updated expected trajectory. At $t=4\,\text{s}$ and $t=6\,\text{s}$, abrupt updates of the desired trajectory cause the drone to deviate significantly.}
    \label{tf0}
\end{figure}

Table \ref{tracking_err} reports RMSE for ten runs across Scenes 1–3. Incorporating $\Delta T_f$ reduces RMSE by preventing abrupt desired‐position updates that degrade tracking.

Fig. \ref{tf0} shows Scene 2 with $\Delta T_f=0\,\text{s}$. Each new trajectory is applied after the drone has moved (see Fig. \ref{tf0-t=4}, \ref{tf0-t=6}), causing large tracking errors. In contrast, by updating the trajectory at a future position relative to the tracking controller's current indexing, the online replanning framework compensates for system delays and avoids abrupt command changes. This explains the lower RMSE observed when $\Delta T_f=1\,\text{s}$ in Table \ref{tracking_err}.

\section{Real-World Experiments}
\begin{figure}[t]
\vspace{7pt}
\setlength{\abovecaptionskip}{\aboveCaptionFig} 
\setlength{\belowcaptionskip}{\belowCaptionFig}
\centerline{\includegraphics[width=1\linewidth]{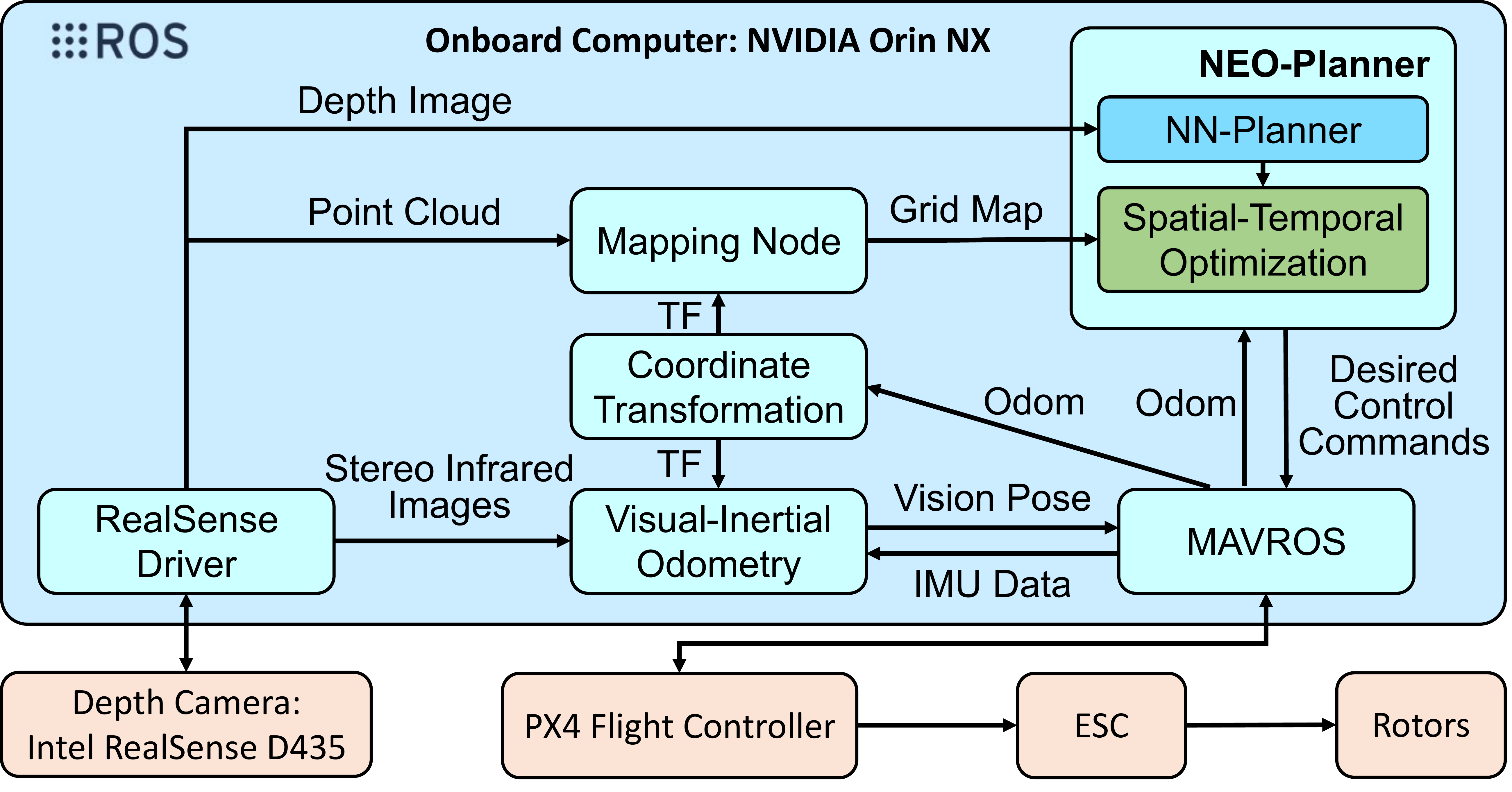}}
\caption{Main software and hardware architecture of the drone.}
\label{drone_system}
\end{figure}

For real-world experiments, we deploy NEO-Planner on a drone equipped with an NVIDIA Jetson Orin NX onboard computer, an Intel RealSense D435 depth camera, and a Holybro Kakute H7 mini flight controller. The drone utilizes VINS-Fusion \cite{qin_vins-mono_2018} for state estimation. The neural network trained in simulations is directly deployed using ONNX-Runtime. NEO-Planner uses the parameters in Table \ref{params}. Fig. \ref{drone_system} illustrates the drone's software and hardware architecture.

\begin{figure}[t]
\vspace{-6pt}
\setlength{\abovecaptionskip}{\aboveCaptionFig} 
\setlength{\belowcaptionskip}{\belowCaptionFig}
\centerline{\includegraphics[width=1\linewidth]{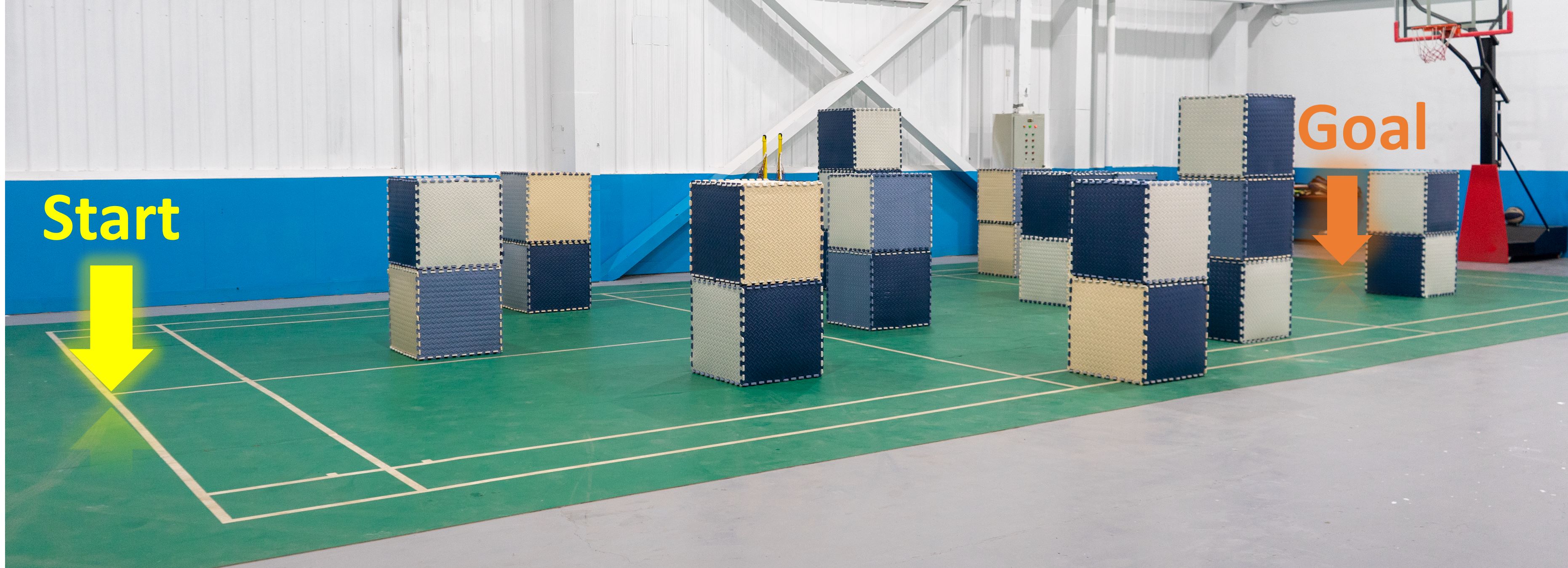}}
\caption{\textbf{The environment for real flight test.} This picture shows the first of the three scenes. The drone takes off from the start point and flies towards the goal point.}
\label{exp_scene}
\end{figure}

\begin{figure}[b]
\vspace{-16pt}
\setlength{\abovecaptionskip}{\aboveCaptionFig} 
\setlength{\belowcaptionskip}{\belowCaptionFig}
\centerline{\includegraphics[width=1\linewidth]{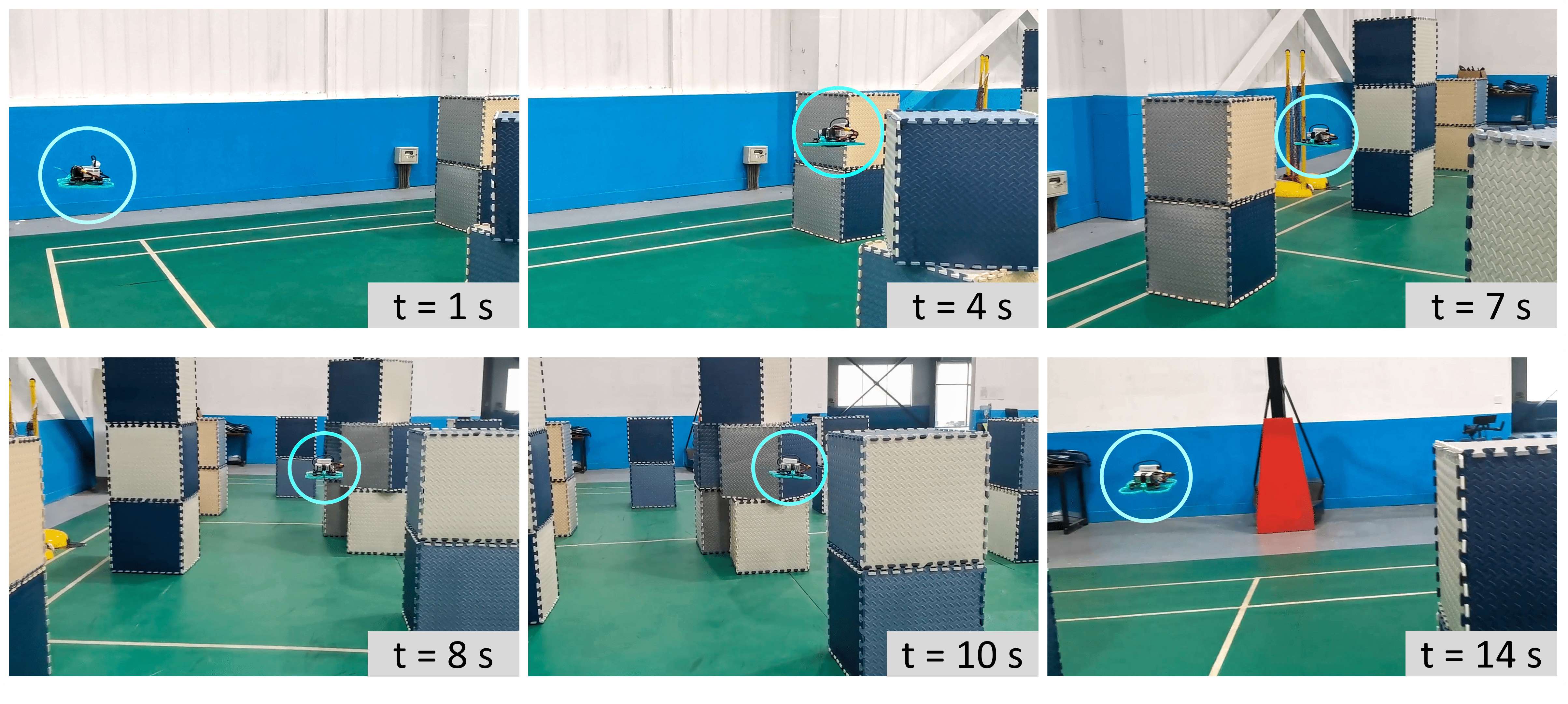}}
\caption{Flight test in real-world Scene 2.}
\label{trial_2}
\end{figure}

\begin{figure}[t]
\vspace{6pt}
\setlength{\abovecaptionskip}{\aboveCaptionFig} 
\setlength{\belowcaptionskip}{\belowCaptionFig}
\centerline{\includegraphics[width=1\linewidth]{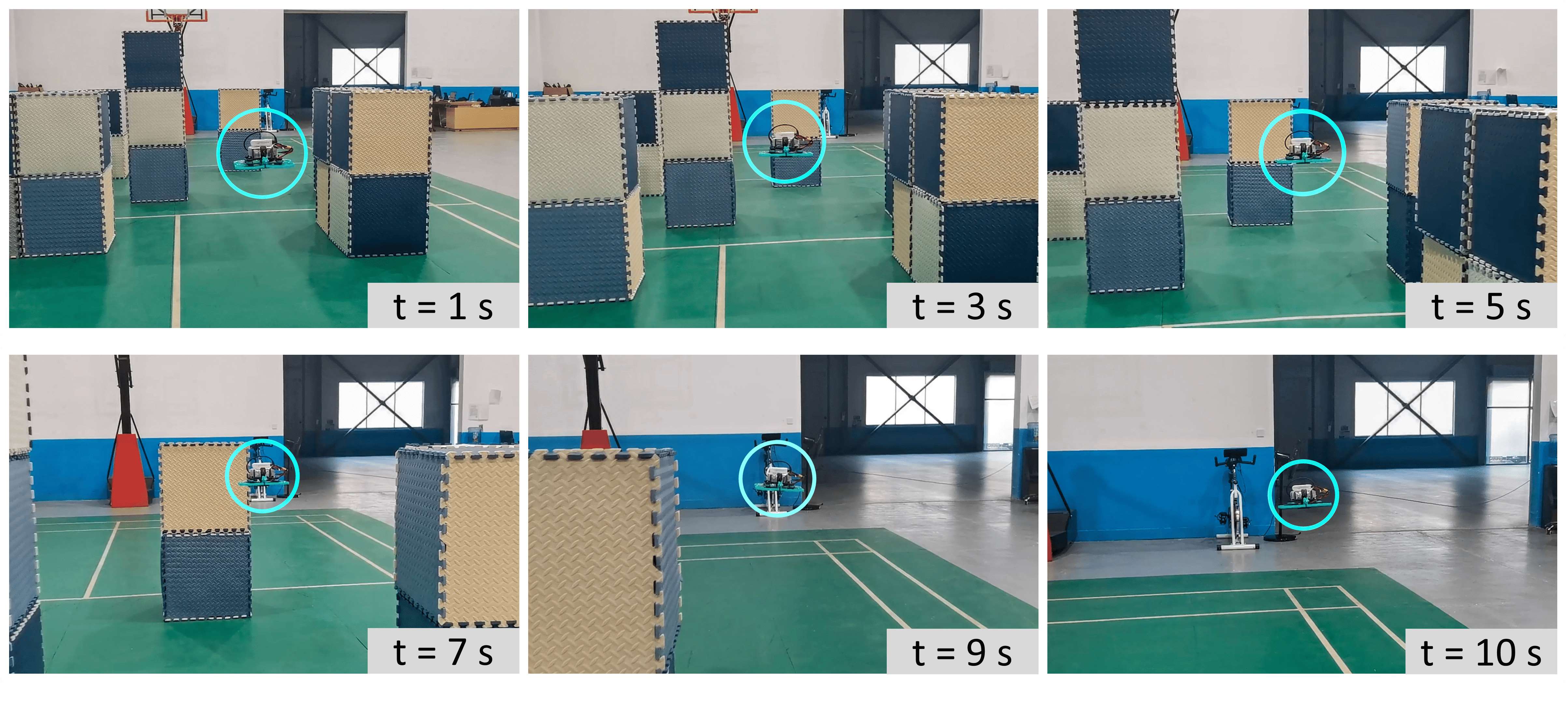}}
\caption{Flight test in real-world Scene 3.}
\label{trial_3}
\end{figure}

We tested in a badminton-court-sized cluttered area (Fig. \ref{exp_scene}) across three obstacle layouts. In each, the drone autonomously took off, traversed the obstacles, and reached the far end. All flights succeeded without collisions (Fig. \ref{exp_process}, Fig. \ref{trial_2}, and Fig. \ref{trial_3}). Recorded data in Table \ref{real_data} show the drone operated near its commanded speed limit of 1.0 m/s., validating NEO-Planner in real-world conditions.

\section{Conclusions}
This paper presented NEO-Planner, a trajectory planner that initializes a trajectory with a neural network and further improves it with spatial-temporal optimization. The neural network's primary role is to provide initial values in both spatial and temporal profiles. Compared to conventional optimization techniques, the incorporation of the neural network has demonstrated substantial reductions in computational time while maintaining a comparable level of trajectory quality. Furthermore, we presented a robust online planning framework that exhibits tolerance towards planning latency and decouples the planning and control frequencies. These methods have been validated through real-world experiments, affirming their practical utility and effectiveness.

\begin{table}[t]
\setlength{\abovecaptionskip}{\aboveCaptionFig} 
\setlength{\belowcaptionskip}{\belowCaptionFig}
\centering
\caption{Real-world flight results in three scenes}
\label{real_data}
\begin{tabular}{cccc}
\toprule
\textbf{Scene} & \textbf{Traj. Length ($\text{m}$)} & \textbf{Travel Time ($\text{s}$)} & \textbf{Avg. Vel. ($\text{m/s}$)} \\ \midrule
1 & 12.51 & 12.00 & 1.04 \\
2 & 16.23 & 14.90 & 1.09 \\
3 & 10.70 & 10.42 & 1.03 \\ \bottomrule
\end{tabular}
\vspace{-10pt}
\end{table}

\bibliographystyle{IEEEtran}
\bibliography{references}

\end{document}